\documentclass{article}

\PassOptionsToPackage{square, numbers, compress,sort}{natbib}
\usepackage[final]{neurips_2018}

\usepackage[utf8]{inputenc} % allow utf-8 input
\usepackage[T1]{fontenc}    % use 8-bit T1 fonts
\usepackage{hyperref}       % hyperlinks
\usepackage{color}
\usepackage{url}            % simple URL typesetting
\usepackage{booktabs}       % professional-quality tables
\usepackage[algo2e,ruled]{algorithm2e}
\usepackage{amsmath,amsfonts,amsthm,amssymb}       % blackboard math symbols
\usepackage{nicefrac}       % compact symbols for 1/2, etc.
\usepackage{microtype}      % microtypography
\usepackage[pdftex]{graphicx}
\usepackage{subcaption}
\usepackage{mwe}
\usepackage{times}
\usepackage{titlesec} % change font heading
\usepackage{multirow}
\usepackage{enumitem}
\usepackage{hhline}  % double horizontal lines for tables
\usepackage{aliascnt} % for using same numbering for theorems and lemmas
\usepackage{wrapfig}

\newtheorem{observation}{Observation}
\newtheorem{theorem}{Theorem}
\newtheorem{lemma}{Lemma}

\newif\ifcomments
\commentstrue
\ifcomments\newcommand{\comments}[1]{#1}\else\newcommand{\comments}[1]{}\fi

\definecolor{clrgp}{rgb}{.9,0,.9}
\definecolor{gray}{rgb}{0.41, 0.41, 0.41}
\definecolor{forestgreen}{rgb}{0.13, 0.55, 0.13}

\titleformat*{\subparagraph}{\normalfont\itshape}
\renewcommand{\paragraph}[1]{\vspace{0.20ex}\noindent\textbf{#1}}
\renewcommand{\subparagraph}[1]{\vspace{0.20ex}\noindent\textit{#1}}

\usepackage{xcolor}
\definecolor{dark-red}{rgb}{0.4,0.15,0.15}
\definecolor{dark-blue}{rgb}{0.15,0.15,0.4}
\definecolor{medium-blue}{rgb}{0,0,0.5}
\hypersetup{
   colorlinks, linkcolor={dark-blue},
   citecolor={dark-blue}, urlcolor={medium-blue}
}

\newcommand{\titl}{GPyTorch: Blackbox Matrix-Matrix Gaussian Process Inference with GPU Acceleration}

\newcommand{\authorinfo}{
  Jacob R. Gardner\thanks{Equal contribution.}, \:
  Geoff Pleiss\footnotemark[1], \\
  {\bf David Bindel,} \:
  {\bf Kilian Q. Weinberger,} \:
  {\bf Andrew Gordon Wilson}
  \\ Cornell University
  \\
  \texttt{\{jrg365,kqw4,andrew\}@cornell.edu},
  \\
  \texttt{\{geoff,bindel\}@cs.cornell.edu}
}
\title{\titl}
\author{\authorinfo}

\DeclareMathOperator*{\argmin}{arg\,min}
\DeclareMathOperator*{\expectedvalue}{\mathbb{E}}

\newcommand{\bigo}[1]{\ensuremath{\mathchoice{\mathcal O \! \left( #1 \right)}}{\mathcal O ( #1 )}{}{}}
\newcommand{\bigomega}[1]{\ensuremath{\mathchoice{\Omega \! \left( #1 \right)}}{\Omega ( #1 )}{}{}}
\newcommand{\reals}{\ensuremath{\mathbb{R}}}

\newcommand{\normaldist}[2]{\ensuremath{\mathcal{N} \left( #1, #2 \right)}}

\newcommand{\Ev}[1]{\ensuremath{\expectedvalue \left[ #1 \right]}}
\newcommand{\Evover}[2]{\ensuremath{\expectedvalue_{#1} \left[ #2 \right]}}

\newif\ifboldmatrix
\boldmatrixfalse
\ifboldmatrix\newcommand{\boldmatrix}[1]{\mathbf{#1}}\else\newcommand{\boldmatrix}[1]{#1}\fi

\newcommand{\x}{\ensuremath{\mathbf{x}}}
\newcommand{\bx}{\ensuremath{\mathbf{x}}}
\newcommand{\ba}{\ensuremath{\mathbf{a}}}
\newcommand{\bb}{\ensuremath{\mathbf{b}}}
\newcommand{\bd}{\ensuremath{\mathbf{d}}}
\newcommand{\by}{\ensuremath{\mathbf{y}}}
\newcommand{\bv}{\ensuremath{\mathbf{v}}}
\newcommand{\bk}{\ensuremath{\mathbf{k}}}

\newcommand{\bq}{\ensuremath{\mathbf{q}}}
\newcommand{\br}{\ensuremath{\mathbf{r}}}

\newcommand{\bw}{\ensuremath{\mathbf{w}}}

\newcommand{\bu}{\ensuremath{\mathbf{u}}}

\newcommand{\X}{\ensuremath{\boldmatrix{X}}}
\newcommand{\M}{\ensuremath{\boldmatrix{M}}}

\newcommand{\B}{\ensuremath{\boldmatrix{B}}}
\newcommand{\A}{\ensuremath{\boldmatrix{A}}}

\newcommand{\D}{\ensuremath{\boldmatrix{D}}}
\newcommand{\Q}{\ensuremath{\boldmatrix{Q}}}
\newcommand{\T}{\ensuremath{\boldmatrix{T}}}

\newcommand{\Z}{\ensuremath{\boldmatrix{Z}}}
\renewcommand{\L}{\ensuremath{\boldmatrix{L}}}
\newcommand{\I}{\ensuremath{\boldmatrix{I}}}
\renewcommand{\P}{\ensuremath{\boldmatrix{P}}}
\newcommand{\U}{\ensuremath{\boldmatrix{U}}}
\newcommand{\V}{\ensuremath{\boldmatrix{V}}}
\newcommand{\K}{\ensuremath{\boldmatrix{K}}}
\newcommand{\y}{\ensuremath{\mathbf{y}}}
\newcommand{\bz}{\ensuremath{\mathbf{z}}}

\newcommand{\dset}{\ensuremath{\mathcal D}}
\newcommand{\eye}{\ensuremath{\boldmatrix{I}}}

\newcommand{\trainK}{\ensuremath{\widehat K_{\X \! \X}}}
\newcommand{\trainP}{\ensuremath{\widehat P}}
\newcommand{\tr}[1]{\ensuremath{\mathchoice{\textrm{Tr} \left( #1 \right)}{\textrm{Tr} ( #1 )}{}{}}}
\newcommand{\mmm}[1]{\ensuremath{\Xi (#1)}}
\newcommand{\mvm}[1]{\ensuremath{\xi ( #1 )}}
\newcommand{\row}[1]{\ensuremath{\rho ( #1 )}}
\newcommand{\mmacro}{BBMM}
\newcommand{\mmfullname}{Blackbox Matrix-Matrix}
\newcommand{\mcgacro}{mBCG}

\newcommand{\R}{\ensuremath{\boldmatrix{R}}}

\begin{document}
\maketitle

\begin{abstract}
  Despite advances in scalable models, the inference tools used for Gaussian processes (GPs) have yet to fully capitalize on developments in computing hardware.
  We present an efficient and general approach to GP inference based on Blackbox Matrix-Matrix multiplication (BBMM).
  BBMM inference uses a modified \emph{batched} version of the conjugate gradients algorithm to derive all terms for training and inference in a single call.
  BBMM reduces the asymptotic complexity of exact GP inference from $\bigo{n^3}$ to $\bigo{n^2}$.
  Adapting this algorithm to scalable approximations and complex GP models simply requires a routine for efficient matrix-matrix multiplication with the kernel and its derivative.
  In addition, BBMM uses a specialized preconditioner to substantially speed up convergence.
  In experiments we show that BBMM effectively uses GPU hardware to dramatically accelerate both exact GP inference and scalable approximations.
  Additionally, we provide \emph{GPyTorch}, a software platform for scalable GP inference via BBMM, built on PyTorch.
\end{abstract}

%!TEX root=main.tex
\section{Introduction}
The past years have witnessed unprecedented innovation in deep learning. This progress has involved innovations in network designs~\cite{alexnet,hahnloser2000digital,he2016deep,huang2017densely,batchnorm}, but it also has benefited vastly from improvements in optimization~\cite{bottou2010large}, and excellent software implementations such as PyTorch, MXNet, TensorFlow and Caffe~\cite{paszke2017automatic,chen2015mxnet,abadi2016tensorflow,jia2014caffe}. Broadly speaking, the gains in optimization originate in large part from insights in stochastic gradient optimization~\citep{bottou2010large, krizhevsky2012imagenet, chaudhari2016entropy, hochreiter1997flat, keskar2016large, izmailov2018averaging}, effectively trading off unnecessary exactness for speed and in some cases regularization. Moreover, the advantages of modern software frameworks for deep learning include rapid prototyping, easy access to specialty compute hardware (such as GPUs), and blackbox optimization through automatic differentiation.

Similarly, Gaussian process research has undergone significant innovations in recent years~\cite{titsias2009variational,hensman2013gaussian,wilson2014thesis,wilson2015kernel,wilson2015thoughts,cunningham2008fast} --- in particular to improve scalability to large data sets. However,
the tools most commonly used for GP inference do not effectively utilize modern hardware, and new models require significant implementation efforts. Often, in fact, the \emph{model} and the \emph{inference engine} are tightly coupled and consequently many complex models like multi-output GPs and scalable GP approximations require custom inference procedures \cite{hensman2015scalable,bonilla2008multi}. This entanglement of model specification and inference procedure impedes rapid prototyping of different model types, and obstructs innovation in the field.

In this paper, we address this gap by introducing a highly efficient framework for Gaussian process inference.
Whereas previous inference approaches require the user to provide routines for computing the full GP marginal log likelihood for a sufficiently complex model,
our framework only requires access to a blackbox routine that performs matrix-matrix multiplications with the kernel matrix and its derivative.
Accordingly, we refer to our method as \mmfullname{} (\mmacro{}) Inference.

In contrast to the Cholesky decomposition, which is at the heart of many existing inference engines, matrix-matrix multiplications fully utilize GPU acceleration.
We will demonstrate that this matrix-matrix approach also significantly eases implementation for a wide class of existing GP models from the literature.
In particular, we make the following contributions:

\noindent
1. Inspired by iterative matrix-vector multiplication (MVM)-based inference methods  \cite{cunningham2008fast,saatcci2012scalable,wilson2015kernel,wilson2015thoughts,dong2017scalable}, we provide a modified \emph{batched} version of linear conjugate gradients (\mcgacro{}) that provides all computations necessary for both the marginal likelihood and its derivatives.
Moreover, \mcgacro{} uses large matrix-matrix multiplications that more efficiently utilize modern hardware than both existing Cholesky and MVM based inference strategies.
Our approach also circumvents several critical space complexity and numerical stability issues present in existing inference methods.
Most notably, \mmacro{} reduces the time complexity of exact GP inference from $\bigo{n^3}$ to $\bigo{n^2}$.

\noindent
2. We introduce a method for \emph{preconditioning} this modified conjugate gradients algorithm based on the pivoted Cholesky decomposition \cite{bach2013sharp,harbrecht2012low}.
All required operations with this preconditioner are efficient, and in practice require negligible time.
We demonstrate both empirically and theoretically that this preconditioner significantly accelerates inference.

\noindent 3. We introduce \textbf{\href{https://gpytorch.ai}{GPyTorch}}, a new software platform using \mmacro{} inference for
scalable Gaussian processes, which is built on top of PyTorch: \url{https://gpytorch.ai}.
On datasets as large as $3000$ data points (until we fill GPU memory) we demonstrate that \emph{exact} GPs with \mmacro{} are \emph{up to $20\times$ faster than GPs using Cholesky-based approaches}.
Moreover, the popular SKI \cite{wilson2015kernel} and SGPR \cite{titsias2009variational} frameworks with \mmacro{} achieve up to $15\times$ and $4\times$ speedups (respectively) on datasets as large as $500,\!000$ data points.
Additionally, SKI, SGPR and other scalable approximations are implemented in \emph{less than 50 lines of code}, requiring only an efficient matrix-matrix multiplication routine.

%!TEX root=main.tex
\section{Related Work}
\paragraph{Conjugate gradients, the Lanczos tridiagonalization algorithm,}
and their relatives are methods from numerical linear algebra for computing linear solves and solving eigenvalue problems \emph{without explicitly computing a matrix}.
These techniques have been around for decades, and are covered in popular books and papers \cite{saad2003iterative,golub2012matrix,demmel1997applied,parlett1980new,lanczos1950iteration,datta2010numerical,paige1970practical}.
These algorithms belong to a broad class of iterative methods known as \emph{Krylov subspace methods}, which access matrices only through matrix-vector multiplies (MVMs).
Historically, these methods have been applied to solving large numerical linear algebra problems, particularly those involving sparse matrices that afford fast MVMs.

Recently, a number of papers have used these MVM methods for parts of GP inference \cite{dong2017scalable,cunningham2008fast,murray2009gaussian,saatcci2012scalable,wilson2014thesis,wilson2015kernel,gardner2018product,pleiss2018constant}.
One key advantage is that MVM approaches can exploit algebraic structure for increased computational efficiencies.
Notably, the structured kernel interpolation (SKI) method \cite{wilson2015kernel} uses structured kernel matrices with fast MVMs to achieve a remarkable asymptotic complexity.
\citet{dong2017scalable} propose MVM methods for computing stochastic estimates of log determinants and their derivatives using a technique based on Lanczos tridiagonalization \cite{golub2009matrices,ubaru2017fast}.
We utilize the same log determinant estimator as \citet{dong2017scalable}, except we avoid explicitly using the Lanczos tridiagonalization algorithm which has storage and numerical stability issues \cite{golub2012matrix}.

\paragraph{Preconditioning} is an effective tool for accelerating the convergence of conjugate gradients.
These techniques are far too numerous to review adequately here; however, \citet{saad2003iterative} contains two chapters discussing a variety of preconditioning techniques.
\citet{cutajar2016preconditioning} explores using preconditioned conjugate gradients for exact GP inference, where they use various sparse GP methods (as well as some classical methods) as preconditioners. However, the methods in \citet{cutajar2016preconditioning} do not provide general purpose preconditioners.
For example, methods like Jacobi preconditioning have no effect when using a stationary kernel \cite{cutajar2016preconditioning,wilson2015thoughts}, and many other preconditioners have $\bigomega{n^{2}}$ complexity, which dominates the complexity of most scalable GP methods.

\paragraph{The Pivoted Cholesky decomposition}
is an efficient algorithm for computing a low-rank decomposition of a positive definite matrix \cite{harbrecht2012low,bach2013sharp}, which we use in the context of preconditioning.
\citet{harbrecht2012low} explores the use of the pivoted Cholesky decomposition as a low rank approximation, although primarily in a scientific computing context.
In proving convergence bounds for our preconditioner we explicitly make use of some theoretical results from \cite{harbrecht2012low} (see \autoref{app:theory}).
\citet{bach2013sharp} considers using random column sampling as well as the pivoted Cholesky decomposition as a low-rank approximation to kernel matrices.
However, \citet{bach2013sharp} treats this decomposition as an approximate training method, whereas we use the pivoted Cholesky decomposition primarily
as a preconditioner, which avoids any loss of accuracy from the low rank approximation as well as the complexity of computing derivatives.

%!TEX root=main.tex
\section{Background}

\paragraph{Notation.}
$\X$ will denote a set of $n$ training examples in $d$ dimensions, or equivalently an $n \times d$ matrix where the $i^{\text{th}}$ row (denoted $\x_{i}$) is the $i^\text{th}$ training example.
$\y$ denotes the training labels.
$k(\x, \x')$ denotes a \emph{kernel function}, and $\K_{\X\X}$ denotes the matrix containing all pairs of kernel entries, i.e. $[\K_{\X\X}]_{ij} = k(\x_{i}, \x_{j})$.
$\bk_{\X\x^{*}}$ denotes kernel values between training examples and a test point $\x^{*}$, e.g. $[\bk_{\X\x^{*}}]_{i} = k(\x_{i}, \x^{*})$.
A hat denotes an added diagonal: $\trainK = \K_{\X\X} + \sigma^{2}I$.

\paragraph{A Gaussian process} (GP) is a kernel method that defines a full distribution over the function being modeled, $f(\bx) \sim \mathcal{GP} \left( \mu(\bx), k(\bx,\bx^{\prime}) \right)$.
Popular kernels include the RBF kernel, $k(\bx, \bx^{\prime}) = s\exp\left(-(\Vert \bx - \bx^{\prime} \Vert)/(2\ell^{2})\right)$ and the Mat\'ern family of kernels \cite{rasmussen2006gaussian}.

\paragraph{Predictions with a Gaussian process.}
Predictions with a GP are made utilizing the \emph{predictive posterior distribution}, $p(f(\bx^{*})\mid\X,\y)$. Given two test inputs $\x^{*}$ and $\x^{*\prime}$, the predictive mean for $\x^{*}$ and the predictive covariance between $\x^{*}$ and $\x^{*\prime}$ are given by:
\begin{align}
  \mu_{f \mid \dset}(\x^*) = \mu(\x^*) + \bk_{\X \x^*}^\top \trainK^{-1} \y,
  &&&&
  k_{f\mid\dset}(\x^*, \x^{*\prime}) = k_{\x^{*} \x^{*\prime}} - \bk_{\X \x^{*}}^\top \trainK^{-1} \bk_{\X \x^{*\prime}},
    \label{eq:pred_mean_covar}
\end{align}
\paragraph{Training a Gaussian process.}
Gaussian processes depend on a number of \emph{hyperparameters} $\theta$. Hyperparameters may include the likelihood noise, kernel lengthscale, inducing point locations \cite{titsias2009variational}, or neural network parameters for deep kernel learning \cite{wilson2016deep}. These parameters are commonly learned by minimization or sampling via the \emph{negative log marginal likelihood}, given (with derivative) by
\begin{equation}
  L(\theta \! \mid \! \X, \y) \propto \log \left\vert \trainK \right\vert - \y^{\top}\trainK^{-1}\y,
  \:\:\:
  \frac{dL}{d\theta} = \y^{\top} \! \trainK^{-1}\frac{d\trainK}{d\theta}\trainK^{-1}\y + \tr{\trainK^{-1}\frac{d\trainK}{d\theta}} \label{eq:log_lik_and_deriv}.
\end{equation}

%!TEX root=main.tex
%\newpage
\section{Gaussian process inference through blackbox matrix multiplication}

\label{sec:method}
The goal of our paper is to replace existing inference strategies with a unified framework that utilizes modern hardware efficiently.
We additionally desire that complex GP models can be used in a blackbox manner without additional inference rules. To this end,
our method reduces the bulk of GP inference to one of the most efficiently-parallelized computations: \emph{matrix-matrix multiplication}.
We call our method \mmfullname{} inference (\mmacro) because it only requires a user to specify a matrix multiply routine for the kernel $\trainK M$ and its derivative $\frac{d\trainK}{d\theta}M$.

\paragraph{Required operations.}
An \emph{inference engine} is a scheme for computing all the equations discussed above: the predictive distribution~\eqref{eq:pred_mean_covar}, the loss, and its derivative~\eqref{eq:log_lik_and_deriv}.
These equations have three operations in common that dominate its time complexity:
1) the linear solve $\trainK^{-1}\y$,
2) the log determinant $\log \vert \trainK \vert$,
and 3) a trace term $\tr{\trainK^{-1}\frac{d\trainK}{d\theta}}$.
In many implementations, these three quantities are computed using the Cholesky decomposition of $\trainK$,
which is computationally expensive, requiring $\bigo{n^3}$ operations, and does not effectively utilize parallel hardware.

Recently, there is a growing line of research that computes these operations with iterative routines based on matrix-vector multiplications (MVMs).
$\trainK^{-1}\y$ can be computed using \emph{conjugate gradients} (CG) \cite{cunningham2008fast,cutajar2016preconditioning,saatcci2012scalable,wilson2015kernel},
and the other two quantities can be computed using calls to the iterative Lanczos tridiagonalization algorithm \cite{ubaru2017fast,dong2017scalable}.
MVM-based methods are asymptotically faster and more space efficient than Cholesky based methods \cite{wilson2015kernel,dong2017scalable}.
Additionally, these methods are able to exploit algebraic structure in the data for further efficiencies \cite{cunningham2008fast,saatcci2012scalable,wilson2015kernel}.
However, they also have disadvantages.
The quantities are computed via several independent calls to the CG and stochastic Lanczos quadrature subroutines, which are inherently sequential and therefore do not fully utilize parallel hardware. Additionally, the Lanczos tridiagonalization algorithm requires $\bigo{np}$ space for $p$ iterations and suffers from numerical stability issues due to loss of orthogonality \cite{golub2012matrix}.

\paragraph{Modified CG.}
Our goal is to capitalize on the advantages of MVM-based methods (space-efficiency, ability to exploit structure, etc.) but with efficient routines that are optimized for modern parallel compute hardware.
For this purpose, our method makes use of a \emph{modified Batched Conjugate Gradients Algorithm} (\mcgacro{}) algorithm.
Standard conjugate gradients takes as input a vector $\y$ and a routine for computing a matrix vector product $\trainK\y$, and, after $p$ iterations, outputs an approximate solve $\bu_{p} \approx \trainK^{-1}\y$ (with exact equality when $p = n$).
We modify conjugate gradients to (1) perform linear solves with multiple right hand sides simultaneously, and (2) return tridiagonal matrices corresponding to partial Lanczos tridiagonalizations of $\trainK$ with respect to each right hand side.\footnote{
  mBCG differes from Block CG algorithms \cite{o1980block} in that mBCG returns Lanczos tridiagonalization terms.
} Specifically, \mcgacro{} takes as input a matrix $\left[\begin{array}{cccc}\by & \bz_{1} & \cdots & \bz_{t}\end{array}\right]$, and outputs:
\begin{equation}
  \label{eq:mod_cg_call}
  \left[\begin{array}{cccc}\bu_{0} & \bu_{1} & \cdots & \bu_{t}\end{array}\right] = \trainK^{-1}\left[\begin{array}{cccc}\by & \bz_{1} & \cdots & \bz_{t}\end{array}\right]\;\;\;\;\textrm{and}\;\;\;\; \tilde{T}_{1},...,\tilde{T}_{t}
\end{equation}
where $\tilde{\T}_{1},\ldots,\tilde{\T}_{t}$ are partial Lanczos tridiagonalizations of $\trainK$ with respect to the vectors $\bz_{1},\ldots,\bz_{t}$, which we describe shortly.
In what follows, we show how to use a single call to \mcgacro{} to compute the three GP inference terms: $\trainK^{-1} \by$, $\tr{ \trainK^{-1} \frac{\partial \trainK}{\partial \theta} }$, and $\log \vert \trainK \vert$.
$\trainK^{-1}\y$ is equal to $\bu_{0}$ in~\eqref{eq:mod_cg_call}, directly returned from \mcgacro{}. We describe the other two terms below.

\paragraph{Estimating $\tr{ \trainK^{-1} \frac{\partial \trainK}{\partial \theta} }$}
from CG relies on \emph{stochastic trace estimation} \cite{avron2011randomized,fitzsimons2016improved,hutchinson1990stochastic}, which allows us to treat this term as a sum of linear solves.
Given i.i.d. random variables $\bz_1, \ldots, \bz_t$ so that $\Ev{\bz_i}=0$ and $\Ev{\bz_i \bz_i^{\top}}=I,
$ (e.g., $\bz_{i} \sim \mathcal{N}(0, I)$)
the matrix trace $\tr{\A}$ can be written as
$
  \tr{\A} = \Ev{\bz_i^\top \A\bz_i}
$
, such that
\begin{equation}
  \label{eq:trace_deriv_estimate}
  \tr{\trainK^{-1}\frac{d\trainK}{d\theta}}  = \Ev{\bz_i^\top  \trainK^{-1}\frac{d\trainK}{d\theta} \bz_i} \approx \frac{1}{t}\sum_{i=1}^{t}\left(\bz_{i}^\top \trainK^{-1}\right)\left(\frac{d\trainK}{d\theta}\bz_{i} \right)
\end{equation}
is an unbiased estimator of the derivative. This computation motivates the $\bz_1, \ldots, \bz_t$ terms in \eqref{eq:mod_cg_call}:
the \mcgacro{} call returns the solves $\trainK^{-1}[\bz_1 \ldots \bz_t]$, which yields $\mathbf{u}_i=\bz_{i}^\top \trainK^{-1}$ . A single matrix multiply with the derivative $\frac{d\trainK}{d\theta}[\bz_1 \ldots \bz_t]$ yields the remaining terms on the RHS. The full trace can then be estimated by elementwise multiplying these terms together and summing, as in
\eqref{eq:trace_deriv_estimate}.

\paragraph{Estimating $\log \vert \trainK \vert$}
can be accomplished using the $\T_{1},...,\T_{t}$ matrices from \mcgacro{}. If $\trainK=\Q\T\Q^{\top}$, with $\Q$ orthonormal, then because $\trainK$ and $\T$ have the same eigenvalues:
\begin{equation}
  \log \vert \trainK \vert = \tr{\log \T}  = \Ev{\bz_i^\top (\log \T)\bz_i} \approx \sum_{i=1}^{t}\bz_{i}^\top \left( \log \T \right) \bz_{i}\label{eq:logdetKQ}
\end{equation}
where $\log \T$ here denotes the matrix logarithm, and the approximation comes from the same stochastic trace estimation technique used for \eqref{eq:trace_deriv_estimate}. One approach to obtain a decomposition $\trainK=\Q\T\Q^{\top}$ is to use the \emph{Lanczos tridiagonalization algorithm}. This algorithm takes the matrix $\trainK$ and a probe vector $\bz$ and outputs the decomposition $\Q\T\Q^{\top}$ (where $\bz$ is the first column of $\Q$). However, rather than running the full algorithm, we can instead run $p$ iterations of the algorithm $t$ times, each with a vector $\bz_{1},...,\bz_{t}$ to obtain $t$ decompositions  $\tilde{\Q}_{1}\tilde{\T}_{1}\tilde{\Q}_{1}^{\top},...,\tilde{\Q}_{t}\tilde{\T}_{t}\tilde{\Q}_{t}^{\top}$ with $\tilde{\Q}_{i} \in \reals^{n \times p}$ and $\tilde{\T}_{i} \in \reals^{p \times p}$. We can use these partial decompositions to estimate~\eqref{eq:logdetKQ}:
\begin{equation}
  \Ev{\bz_i^\top (\log \T)\bz_i} = \Ev{\bz_i^\top  \tilde{\Q}_{i}(\log \tilde{\T}_{i}) \tilde{\Q}_{i}^{\top}\bz_{i}} \approx \frac{1}{t} \sum_{i=1}^{t} \bz_i^{\top} \tilde{\Q}_{i}(\log \tilde{\T}_{i})\tilde{\Q}_{i}^{\top}\bz_{i} = \frac{1}{t} \sum_{i=1}^{t}e_{1}^{\top}(\log \tilde{\T}_{i})e_{1},
  \label{eq:slq}
\end{equation}
where $e_{1}$ is the first row of the identity matrix. Running Lanczos with a starting vector $\bz_{i}$ ensures that all columns of $\tilde{\Q}_{i}$ are orthogonal to $\bz_{i}$ except the first, so $\tilde{\Q}_{i}\bz_{i} = e_{1}$ \cite{dong2017scalable,ubaru2017fast,golub2009matrices}.

In \mcgacro{}, we adapt a technique from \citet{saad2003iterative} which allows us to compute $\tilde{\T}_{1},\ldots,\tilde{\T}_{t}$ corresponding to the input vectors $\bz_{1},\ldots,\bz_{t}$  to \mcgacro{} from the coefficients of CG in $\bigo{1}$ additional work per iteration.
This approach allows us to compute a log determinant estimate identical to \eqref{eq:slq} \emph{without running the Lanczos algorithm}.
Thus we avoid the extra computation, storage, and numerical instability associated with Lanczos iterations.
We describe the details of this adaptation in \autoref{app:mod_cg}.

\paragraph{Runtime and space.}
As shown above, we are able to approximate all inference terms from a single call to \mcgacro{}.
These approximations improve with the number of \mcgacro{} iterations.
Each iteration requires one matrix-matrix multiply with $\trainK$, and the subsequent work to derive these inference terms takes negligible additional time (\autoref{app:method_runtime}).
Therefore, $p$ iterations of \mcgacro{} requires $\bigo{nt}$ space (see \autoref{app:method_runtime}) and $\bigo{p \: \mmm{\trainK}}$ time,
where $\mmm{\trainK}$ is the time to multiply $\trainK$ by a $n \times t$ matrix.
This multiplication takes $\bigo{n^2 t}$ time with a standard matrix.
It is worth noting that this is a lower asymptotic complexity that standard Cholesky-based inference, which is $\bigo{n^3}$.
Therefore, \mmacro{} offers a computational speedup for exact GP inference.
As we will show in \autoref{sec:advantages}, this time complexity can be further reduced with structured data or sparse GP approximations.
\subsection{Preconditioning}
\label{sec:preconditioning}
While each iteration of \mcgacro{} performs large parallel operations that utilize hardware efficiently, the iterations themselves are sequential.
A natural goal for better utilizing hardware is to trade off fewer sequential steps for slightly more effort per step.
We accomplish this goal using \emph{preconditioning} \cite{golub2012matrix,saad2003iterative,demmel1997applied,van2003iterative}, which introduces a positive definite matrix $\P$ to solve the related linear system
\begin{equation*}
  \left( \P^{-1/2} \trainK \P^{-1/2} \right) \P^{1/2} \bu = \P^{-1/2}\y
\end{equation*}
instead of $\trainK^{-1} \y$.
Both systems are guaranteed to have the same solution, but the preconditioned system's convergence depends on the conditioning of $\P^{-1/2} \trainK \P^{-1/2}$ rather than that of $\trainK$.
Crucially, the mBCG and standard CG algorithms only require access to $\P^{-1}$ and not its square root
(see \autoref{alg:std_pcg} and \autoref{alg:mod_pcg}).

\paragraph{Preconditioned BBMM.}
We have to make adjustments to BBMM algorithm in order to use preconditioning.
%While the preconditioned system simply returns $\trainK^{-1} \by$, we need to modify the random probe vectors $\bz_{1}$, $\ldots$, $\bz_{T}$ to get correct estimates of $\log \vert \trainK \vert$ and $\tr{ \trainK_{-1}  \partial \trainK / \partial \theta }$.
Instead of performing solves against probe vectors with unit covariance, the input to the {preconditioned} mBCG algorithm is
\begin{equation}
  \label{eqn:mod_cg_call_precond}
  \left[ \by, \:\: \bz_{1}, \:\: \cdots, \:\: \bz_{T} \right], \quad \bz_{i} \sim \normaldist{0}{\P},
\end{equation}
which produces the solves
$\trainK^{-1} \left[ \by, \:\: \bz_{1}, \:\: \cdots, \:\: \bz_{T} \right], \quad \bz_{i} \sim \normaldist{0}{\P}$.
To understand why this is the case, recall that our log determinant estimate is given by
$
	\Evover{\bz_{i} \sim \normaldist{0}{\I} }{\bz_{i^\top} \Q_{i} \left( \log \T_{i} \right) \Q_{i^\top} \bz_{i}}.
$
If we precondition mBCG with $\P$, then the $\T_{i}$ matrices will correspond to the \emph{preconditioned system}
$(\P^{-1/2} \trainK \P^{-1/2})$ and the $\bz_i$ vectors will also be preconditioned with $\P^{-1/2}$.
Consequentially, the stochastic Lanczos quadrature estimate will return
\begin{align}
	\log \left\vert \P^{- \frac 1 2} \trainK \P^{- \frac 1 2} \right\vert
	&\approx \!
	\Evover{\bz_{i} \sim \normaldist{0}{\P} }{
		\left( \bz_{i^\top} \P^{-\frac 1 2} \right) \!
    \Q_{i} \! \left( \log \T_{i} \right) \! \Q_{i}^\top
		\!\! \left( \P^{-\frac 1 2} \bz_{i} \right)
	}.
	\label{eqn:slq_precond}
\end{align}
By using $\bz_{i} \sim \normaldist{0}{\P}$ as probe vectors:
the resulting preconditioned vectors $\P^{-1/2} \bz_{i}$ will be samples from $\normaldist{0}{\I}$, which is the requirement for a stochastic trace estimate.

We can recover $\log \vert \trainK \vert$ from \eqref{eqn:slq_precond} using the identity
$
  \log \vert \trainK \vert = \log \vert \P^{- 1/2} \trainK \P^{- 1/2} \vert + \log \vert \P \vert.
$
To estimate $\tr{ \trainK^{-1} (d \trainK / d \theta) }$ using the $\bz_{i} \sim \normaldist{0}{\P}$ probe vectors,
we note that we can form a stochastic trace estimate from the following:
\begin{align}
	\tr{\trainK^{-1}\frac{d \trainK}{d \theta}}
	&=
	\tr{
		\trainK^{-1}\frac{d \trainK}{d \theta}
		\Evover{\bz_{i} \sim \normaldist{0}{\P} }{
			\P^{-1} \bz_{i} \bz_{i^\top}
		}
	}
  \nonumber \\
	&\approx
	\Evover{\bz_{i} \sim \normaldist{0}{\P} }{
		\left( \bz_{i} \trainK^{-1} \right)
		\left( \frac{d \trainK}{d \theta} \: \P^{-1} \bz_{i} \right)
	}.
	\label{eqn:trace_est_precond}
\end{align}
The only difference between \eqref{eqn:trace_est_precond} and the non-preconditioned trace estimate in \eqref{eq:trace_deriv_estimate} are
the derivative term $d \trainK / d \theta$ is applied to the preconditioned vectors $\P^{-1} \bz_{i}$.

\paragraph{Requirements of BBMM preconditioners.}
Based on the above discussion, we observe three requirements of any preconditioner $\P$ for BBMM.
First, in order to ensure that preconditioning operations do not dominate the running time of \autoref{alg:mod_pcg}, the preconditioner should afford roughly linear-time solves and linear space.
Second, we should be able to efficiently compute the log determinant of the preconditioner $\log \vert \P \vert$ to ``correct'' the log determinant estimate in \eqref{eqn:slq_precond}.
Finally, we should be able to efficiently sample probe vectors $\bz_{i}$ from the distribution $\normaldist{0}{\P}$.

%We observe two requirements of a preconditioner to be used in general for GP inference.
%First, in order to ensure that preconditioning operations do not dominate running time when using scalable GP methods, the preconditioner should afford roughly linear time solves and space.
%Second, we should be able to efficiently compute the log determinant of the preconditioner matrix, $\log \vert P \vert$.
%This is because the \mcgacro{} algorithm applied to the preconditioned system estimates $\log \vert \P^{-1}\trainK \vert$ rather than $\log \vert \trainK \vert$. We must therefore compute
%$
  %\label{eq:logdet_adjusted}
  %\log \vert \trainK \vert = \log \vert \P^{-1}\trainK \vert + \log \vert P \vert.
%$

\paragraph{The Pivoted Cholesky Decomposition.}
For one possible preconditioner, we turn to the \emph{pivoted Cholesky} decomposition.
The pivoted Cholesky algorithm allows us to compute a low-rank approximation of a positive definite matrix, $\K_{\X\X} \approx L_{k}L_{k}^{\top}$ \cite{harbrecht2012low}.
We precondition mBCG with $(L_k L_k^\top + \sigma^2 \I)^{-1}$, where $\sigma^2$ is the Gaussian likelihood's noise term.
Intuitively, if $\P_{k}=\L_{k}\L_{k}^{\top}$ is a good approximation of $\K_{\X\X}$, then $(\P_{k} + \sigma^{2}\I)^{-1/2}\trainK (\P_{k} + \sigma^{2}\I)^{-1/2} \approx \I$.

While we review the pivoted Cholesky algorithm fully in \autoref{app:pivoted_cholesky}, we would like to emphasize three key properties. First, it can be computed in $\bigo{\row{\K_{\X\X}}k^{2}}$ time, where $\row{\K_{\X\X}}$ is the time to access a row (nominally this is $\bigo{n}$).
Second, linear solves with $\trainP = \L_{k}\L_{k}^{\top} + \sigma^{2}\I$ can be performed in $\bigo{nk^{2}}$ time.
Third, sampling from $\normaldist{0}{\P}$ takes $\bigo{nk}$ time using the reparameterization trick \citep{kingma2014auto}.
Finally, the log determinant of $\trainP$ can be computed in $\bigo{nk^{2}}$ time.
In \autoref{sec:results} we empirically show that this preconditioner dramatically accelerates CG convergence.
Further, in \autoref{app:theory}, we prove the following lemma and theorem for univariate RBF kernels:

\begin{lemma}
  \label{thm:condition_number}
  Let $\K_{\X\X} \in \reals^{n \times n}$ be a univariate RBF kernel matrix.
  Let $\L_{k} \L_k^\top$ be the rank $k$ pivoted Cholesky decomposition of $\K_{\X\X}$, and let $\trainP_{k} = \L_k \L_k^\top + \sigma^{2}\I$.
  Then there exists a constant $b>0$ so that the condition number $\kappa(\trainP^{-1}\trainK)$ satisfies the following inequality:
  \begin{align}
    \kappa \left( \trainP_{k}^{-1/2}\trainK\trainP_{k}^{-1/2} \right) =
    \kappa \left( \trainP_{k}^{-1}\trainK \right)
    &\triangleq \left\Vert \trainP_{k}^{-1}\trainK \right\Vert_{2} \left\Vert \trainK^{-1}\trainP_{k} \right\Vert_{2}
    \nonumber
    \\
    &\leq \left( 1 + \bigo{n\exp(-bk)} \right)^2.
  \end{align}
\end{lemma}
\begin{theorem}[Convergence of pivoted Cholesky-preconditioned CG]
  \label{thm:cg_convergence_rbf}
  Let $\K_{\X\X} \in \reals^{n \times n}$ be a $n \times n$ univariate RBF kernel, and let $\L_k \L_k^\top$ be its rank $k$ pivoted Cholesky decomposition.
  Assume we are using preconditioned CG to solve the system $\trainK^{-1} \y = (\K_{\X\X} + \sigma^2 \I)^{-1} \y$ with preconditioner $\trainP = (\L_k \L_k^\top + \sigma^2 \I)$.
  Let $\bu_p$ be the $p^\textrm{th}$ solution of CG, and let $\bu^{*} = \trainK^{-1} \y$ be the exact solution.
  Then there exists some $b > 0$ such that:
  \begin{equation}
    \Vert \bu^{*} - \bu_{p} \Vert_{\trainK}
    \leq 2 \left(1/(1 + \bigo{\exp(kb)/n}\right)^{p} \left\Vert \bu^{*} - \bu_{0} \right\Vert_{\trainK}.
  \end{equation}
\end{theorem}
\autoref{thm:cg_convergence_rbf} implies that we should expect the convergence of conjugate gradients to improve \emph{exponentially} with the rank of the pivoted Cholesky decomposition used for RBF kernels. In our experiments we observe significantly improved convergence for other kernels as well (\autoref{sec:results}). Furthermore, we can leverage \autoref{thm:condition_number} and existing theory from \cite{ubaru2017fast} to argue that preconditioning improves our log determinant estimate. In particular, we restate Theorem 4.1 of \citet{ubaru2017fast} here:
\begin{theorem}[Theorem 4.1 of \citet{ubaru2017fast}]
  \label{thm:slq_convergence}
  Let $\K_{\X\X} \in \reals^{n \times n}$, and let $\L_k \L_k^\top$ be its rank $k$ pivoted Cholesky decomposition.
  Suppose we run $p \geq \frac{1}{4} \sqrt{ \kappa \left( \trainP_{k}^{-1}\trainK \right) } \log \frac{D}{\epsilon}$ iterations of \mcgacro{},
  where $D$ is a term involving this same condition number that vanishes as $k \to n$ (see \cite{ubaru2017fast}),
  and we use $t \geq \frac{32}{\epsilon^{2}}\log(2/\delta)$ vectors $\bz_{i}$ for the solves.\footnote{
    Our constant of $32/\epsilon^2$ differs from the $24/\epsilon^2$ constant used by \citet{ubaru2017fast}.
    This is because we use Gaussian random probe vectors for $\bz_i$, whereas \citeauthor{ubaru2017fast} use Rademacher probe vectors.
    See \citep[][Table~1]{avron2011randomized} for more details on this constant.
  } Let $\Gamma$ be the log determinant estimate from \eqref{eq:slq}. Then:
  \begin{equation}
    \textrm{Pr}\left[\vert \log \vert \trainP^{-1}\trainK \vert - \Gamma \vert \leq \epsilon\vert \log \vert \trainP^{-1}\trainK \vert \vert \right] \geq 1 - \delta.
  \end{equation}
\end{theorem}
Because \autoref{thm:condition_number} states that the condition number $\kappa \left( \trainP_{k}^{-1}\trainK \right)$ decays exponentially with the rank of $\L_{k}$, \autoref{thm:slq_convergence} implies that we should expect that the number of CG iterations required to accurately estimate $\log \vert \trainP^{-1}\trainK \vert$ decreases quickly as $k$ increases.
In addition, in the limit as $k \rightarrow n$ we have that $\log \vert \trainK \vert = \log \vert \trainP \vert$.
This is because $\log \vert \trainP^{-1}\trainK \vert \rightarrow 0$ (since $\trainP^{-1}\trainK$ converges to $\I$) and we have that $\log \vert \trainK \vert = \log \vert \trainP^{-1}\trainK \vert + \log \vert \trainP \vert$.
Since our calculation of $\log \vert \trainP \vert$ is exact, our final estimate of $\log \vert \trainK \vert$ becomes more exact as $k$ increases.
In future work we hope to derive a more general result that covers multivariate settings and other kernels.
\section{Programmability with \mmacro{}}
\label{sec:advantages}
We have discussed how the \mmacro{} framework is more hardware efficient than existing inference engines, and avoids numerical instabilities with Lanczos. Another key advantage of \mmacro{} is that it can easily be adapted to complex GP models or structured GP approximations.

Indeed \mmacro{} is \emph{blackbox} by nature, only requiring a routine to perform matrix-multiplications with the kernel matrix and its derivative.
Here we provide examples of how existing GP models and scalable approximations can be easily implemented in this framework.
The matrix-multiplication routines for the models require at most \emph{50 lines of Python code}.
All our software, including the following GP implementations with \mmacro{}, are available through our GPyTorch library: \\
\url{https://gpytorch.ai}.

\paragraph{Bayesian linear regression} can be viewed as GP regression with the special kernel matrix $\trainK = \X\X^{\top} + \sigma^{2}I$.
A matrix multiply with this kernel against an $n \times t$ matrix $\V$, $(\X\X^{\top} + \sigma^{2}I)\V$ requires $\bigo{tnd}$ time.
Therefore, \mmacro{} requires $\bigo{ptnd}$ time, and is exact in $\bigo{tnd^2}$ time.
This running time complexity matches existing efficient algorithms for Bayesian linear regression, \emph{with no additional derivation}.
Multi-task Gaussian processes \cite{bonilla2008multi} can be adapted in the same fashion \cite{gardner2018product}.

\paragraph{Sparse Gaussian Process Regression (SGPR)} \cite{titsias2009variational} and many other sparse GP techniques \cite{quinonero2005unifying,snelson2006sparse,hensman2013gaussian} use the subset of regressors (SoR) approximation for the kernel:
$
  \trainK \approx (K_{XU}K_{UU}^{-1}K_{UX} + \sigma^{2}I).
$
Performing a matrix-matrix multiply with this matrix requires $\bigo{tnm + tm^{3}}$ time by distributing the vector multiply and grouping terms correctly.
This computation is \emph{asymptotically faster} than the $\bigo{nm^{2} + m^{3}}$ time required by Cholesky based inference. Augmenting the SoR approximation with a diagonal correction, e.g. as in FITC \cite{snelson2006sparse}, is similarly straightforward.

\paragraph{Structured Kernel Interpolation (SKI)} \cite{wilson2015kernel}, also known as KISS-GP, is an inducing point method designed to provide fast matrix vector multiplies (MVMs) for use with Krylov subspace methods. SKI is thus a natural candidate for \mmacro{} and can benefit greatly from hardware acceleration.
SKI is a generalization of SoR, which specifies $\K_{XU} \approx W\K_{UU}$, where $W$ is a sparse matrix. For example $W$ can correspond to the coefficients of sparse local cubic convolution interpolation.
The SKI approximation applied to the training covariance matrix gives us
$
\trainK \approx (WK_{UU}W^{\top} \! + \! \sigma^{2}I).
$
Assuming no structure in $\K_{UU}$ a matrix multiply requires $\bigo{tn \! + \! tm^{2}}$ time. In KISS-GP \citep{wilson2015kernel,wilson2015thoughts}, the matrix $\K_{UU}$ is also chosen to have algebraic structure, such as Kronecker or Toeplitz structure, which further accelerates MVMs. For example, MVMs with a Toeplitz $\K_{UU}$ only require $\bigo{m \log m}$ time. Thus KISS-GP
provides $\bigo{tn \! + \! tm \log m}$ matrix-matrix multiplies \cite{wilson2015kernel}.

\paragraph{Compositions of kernels} can often be handled automatically.
For example, given a \mmacro{} routine for $\K_{1},\K_{2},\K_{3}$, we can automatically perform $(\K_{1}\K_{2}+\K_{3})M = \K_{1}(\K_{2}M) + \K_{3}M$.
SGPR and KISS-GP are implemented in this fashion. Given some pre-defined basic compositionality strategies, the kernel matrix multiplication $\K\M$ in SGPR reduces to defining how to perform $\K_{\U\U}^{-1}M$, and similarly for KISS-GP it reduces to performing multiplication with a Toeplitz matrix $\K_{\U\U}M$. For product kernels one can follow Gardner et al.~\cite{gardner2018product}.

%!TEX root=main.tex
\section{Results}

We evaluate the \mmacro{} framework, demonstrating: (1) the \mmacro{} inference engine provides a substantial speed benefit over Cholesky based inference and standard MVM-based CG inference, especially for GPU computing; (2) \mmacro{} achieves comparable or better final test error compared to Cholesky inference, even with no kernel approximations; and (3) preconditioning provides a substantial improvement in the efficiency of our approach.

\paragraph{Baseline methods.} We test \mmacro{} on three types of GPs:
1. {\bf Exact} GP models,
2. {\bf SGPR} inducing point models \cite{titsias2009variational,hensman2013gaussian},
and 3. {\bf SKI} models with Toeplitz $\K_{UU}$ and deep kernels \cite{wilson2015kernel,wilson2016deep}.
For Exact and SGPR, we compare \mmacro{} against Cholesky-based inference engines implemented in GPFlow \cite{matthews2017gpflow}.
GPFlow is presently the fastest implementation of these models with a Cholesky inference engine.
Since SKI is not intended for Cholesky inference, we compare \mmacro{} to the inference procedure of \citet{dong2017scalable}, implemented in our GPyTorch package.
This procedure differers from \mmacro{} in that it computes $\trainK^{-1} \y$ without a preconditioner and computes $\log \vert \trainK \vert$ and its derivative with the Lanczos algorithm.

\paragraph{Datasets.}
We test Exact models on five datasets from the UCI dataset repository \cite{asuncion2007uci} with up to 3500 training examples (the largest possible before all implementations exhausted GPU memory): Skillcraft, Gas, Airfoil, Autompg, and Wine.
We test SGPR on larger datasets ($n$ up to 50000): KEGG, Protein, Elevators, Kin40k, and PoleTele.
For SKI we test five of the largest UCI datasets ($n$ up to 515000): Song, Buzz, Protein, Kin40k, and KEGG.

\begin{figure}[t]
  \includegraphics[width=\textwidth]{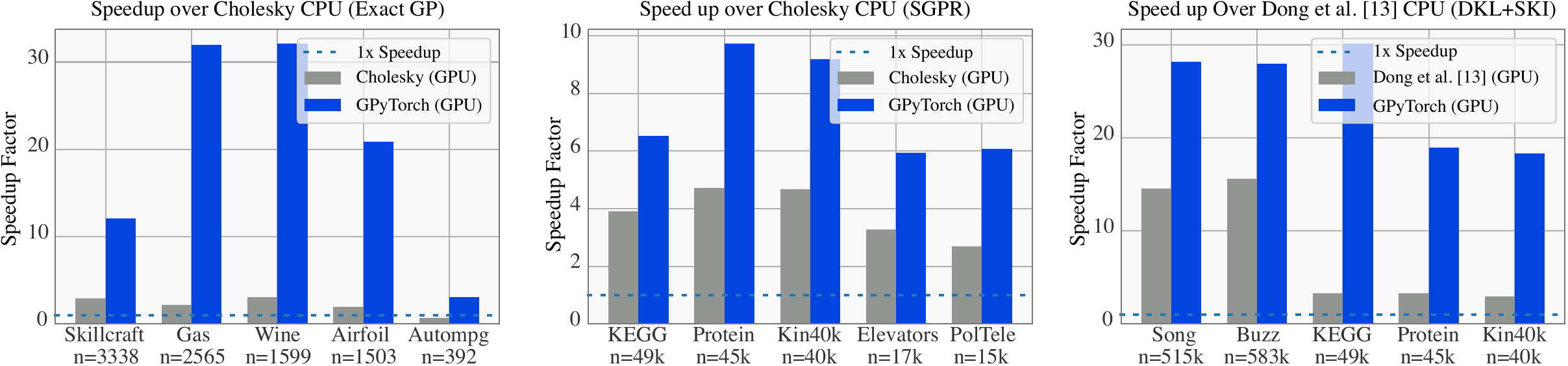}
  \caption{
    Speedup of GPU-accelerated inference engines.
    \mmacro{} is in blue, and competing GPU methods are in gray.
    {\bf Left:} Exact GPs. {\bf Middle:} SGPR \cite{titsias2009variational,hensman2013gaussian} -- speedup over CPU Cholesky-based inference engines.
    {\bf Right:} SKI+DKL \cite{wilson2015kernel,wilson2016deep} -- speedup over CPU inference of \citet{dong2017scalable}.
  }
  \label{fig:timing_results}
\end{figure}

\paragraph{Experiment details.} All methods use the same optimizer (Adam) with identical hyperparameters.
In \mmacro{} experiments we use rank $k\!=\!5$ pivoted Cholesky preconditioners unless otherwise stated.
We use a maximum of $p\!=\!20$ iterations of CG for each solve, and
we use $t\!=\!10$ probe vectors filled with Rademacher random variables to estimate the log determinant and trace terms.
SGPR models use $300$ inducing points.
SKI models use $10,\!000$ inducing points and  the deep kernels described in \cite{wilson2016deep}.
The \mmacro{} inference engine is implemented in our GPyTorch package.
\begin{wrapfigure}{r}{0.5\textwidth}
  \begin{center}
    \includegraphics[width=0.48\textwidth]{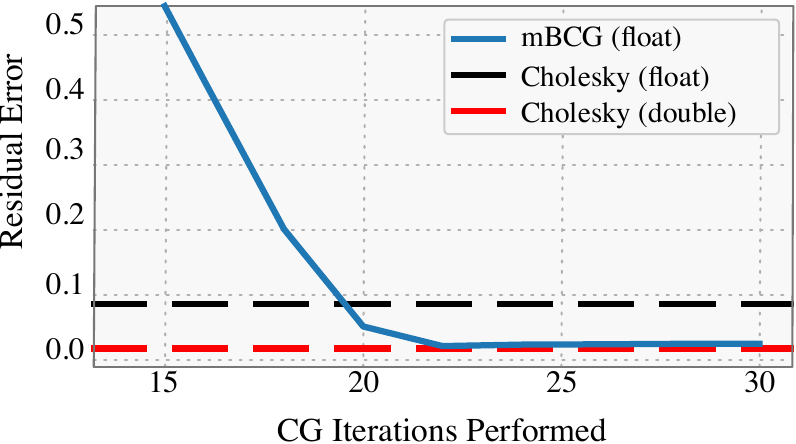}
  \end{center}
  \caption{Solve error for mBCG and Cholesky. \label{fig:cg_error}}
\end{wrapfigure}
All speed experiments are run on an Intel Xeon E5-2650 CPU and an NVIDIA Titan Xp GPU.

\label{sec:results}

\paragraph{Speed comparison.}
\autoref{fig:timing_results} shows the speedup obtained by GPU-accelerated \mmacro{} over the leading CPU-based inference engines (Cholesky for Exact/SGPR, \citet{dong2017scalable} for SKI).
As would be expected, GPU-accelerated \mmacro{} is faster than CPU-based inference.
On Exact and SKI, \mmacro{} is up to \emph{32 times faster} than CPU inference, and up to 10 times faster on SGPR.
The largest speedups occur on the largest datasets, since smaller datasets experience larger GPU overhead.
Notably, \mmacro{} achieves a much larger speedup than GPU accelerated Cholesky methods (Exact, SGPR), which only achieve a roughly $4\times$ speedup.
This result underscores the fact that Cholesky methods are not as well suited for GPU acceleration.
Additionally, \mmacro{} performs better than the GPU-accelerated version of \cite{dong2017scalable} on SKI.
This speedup is because \mmacro{} is able to calculate all inference terms in parallel, while \cite{dong2017scalable} computes the terms in series.

\begin{figure}[t]
  \centering
  \includegraphics[width=\textwidth]{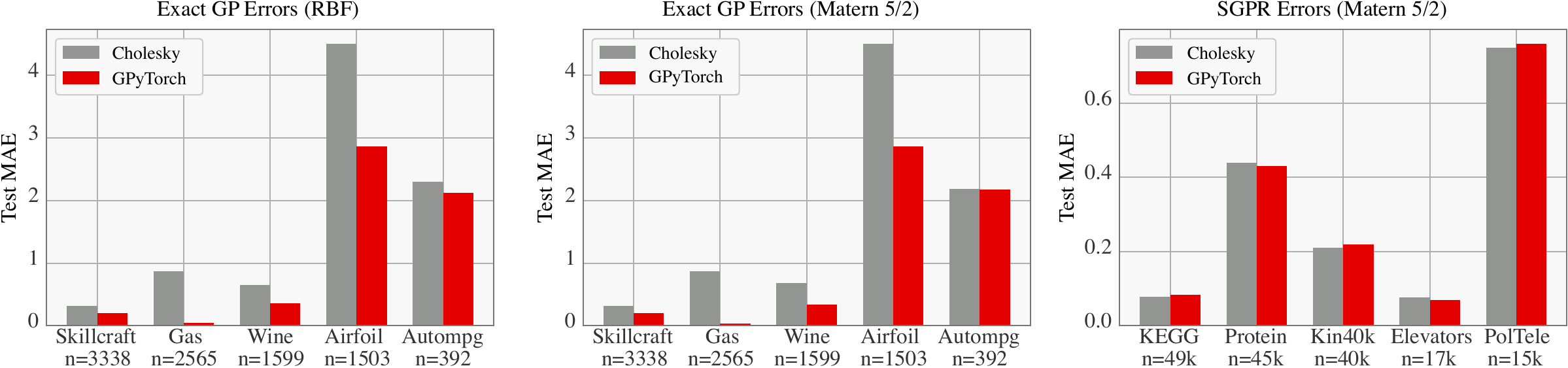}
  \caption{Comparing final Test MAE when using \mmacro{} versus Cholesky based inference. The left two plots compare errors using Exact GPs with RBF and Matern-5/2 kernels, and the final plot compares error using SGPR with a Matern-5/2 kernel on significantly larger datasets.}
  \label{fig:error_results}
\end{figure}

\paragraph{Error comparison.}
In \autoref{fig:error_results} we report test mean average error (MAE) for Exact and SGPR models.\footnote{
  SKI models are excluded from \autoref{fig:error_results}.
  This is because the \mmacro{} inference engine and the inference engine of \citet{dong2017scalable} return identical outputs (see \autoref{app:mod_cg}) even though \mmacro{} is faster.
}
We demonstrate results using both the RBF kernel and a Matern-5/2 kernel.
Across all datasets, our method is at least as accurate in terms of final test MAE.
On a few datasets (e.g. Gas, Airfoil, and Wine with Exact GPs) \mmacro{} even improves final test error.
CG has a regularizing effects which may improve methods involving the exact kernel over the Cholesky decomposition, where numerical issues resulting from extremely small eigenvalues of the kernel matrix are ignored.
For example, Cholesky methods frequently add noise (or ``jitter'') to the diagonal of the kernel matrix for numerical stability.
It is possible to reduce the numerical instabilities with double precision (see \autoref{fig:cg_error}); however, this requires an increased amount of computation.
\mmacro{} on the other hand avoids adding this noise, without requiring double precision.

\begin{figure}[t]
  \centering
  \includegraphics[width=\textwidth]{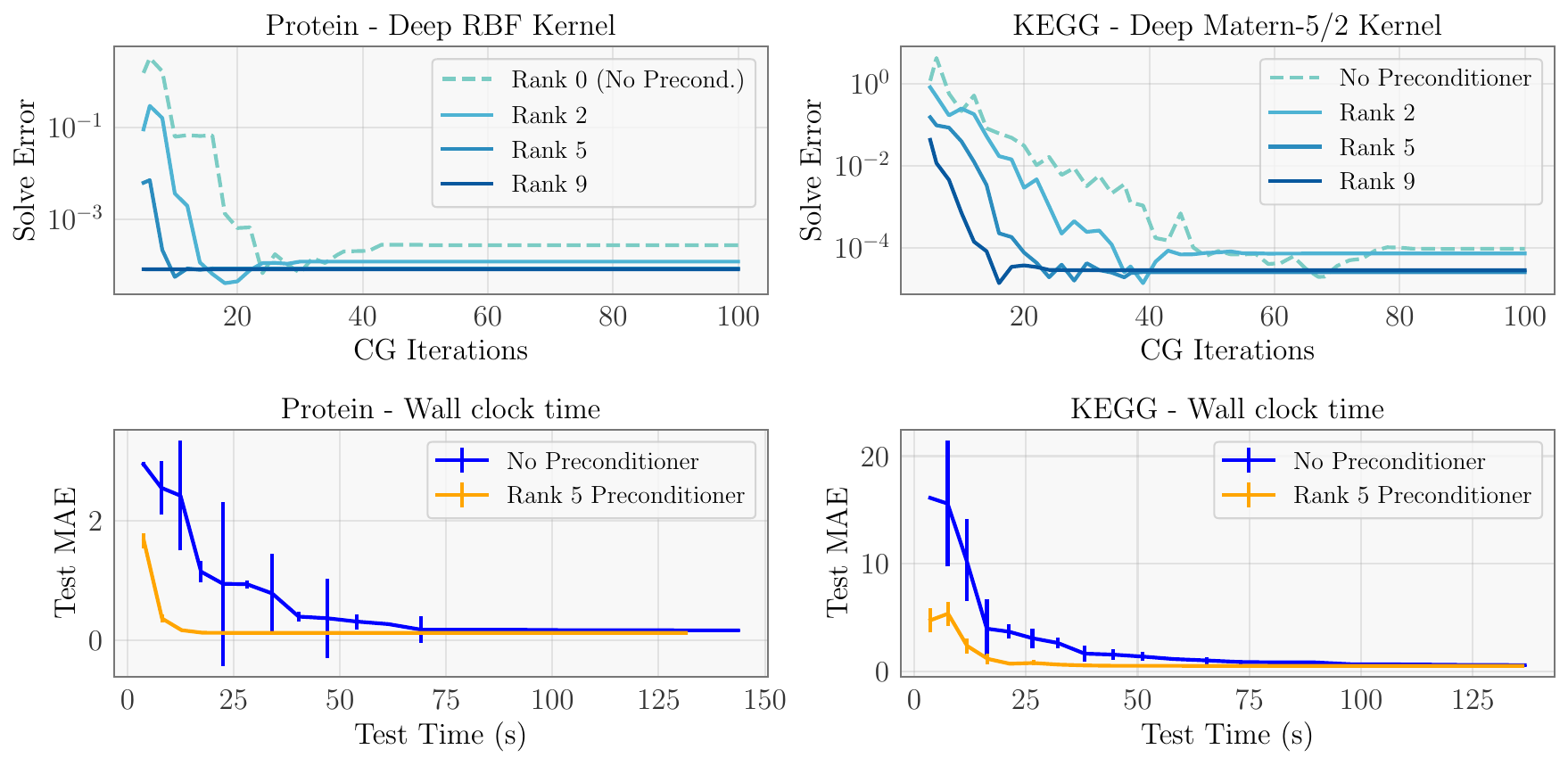}
  \caption{
    The effect of preconditioning on solve errors $\Vert K\x^{*} - \by \Vert / \Vert \by \Vert$ achieved by linear conjugate gradients using no preconditioner versus rank 2, 5, and 9 pivoted Cholesky preconditioners on 2 UCI benchmark datasets using deep RBF and deep Matern kernels.
    The hyperparameters of $K$ were learned by maximizing the marginal log likelihood on each dataset.
  }
  \label{fig:precond_results}
\end{figure}

\paragraph{Preconditioning.}
To demonstrate the effectiveness of preconditioning at accelerating the convergence of conjugate gradients, we first train a deep RBF kernel model on two datasets, Protein and KEGG, and evaluate the solve error of performing $\trainK^{-1}\y$ in terms of the relative residual $\Vert \trainK\bu - \y \Vert / \Vert \y \Vert$ as a function of the number of CG iterations performed.
We look at this error when using no preconditioner, as well as a rank 2, 5, and 9 preconditioner.
To demonstrate that the preconditioner is not restricted to use with an RBF kernel, we evaluate using a deep RBF kernel on Protein and a deep Matern-5/2 kernel on KEGG.
The results are in the top of \autoref{fig:precond_results}.
As expected based on our theoretical intuitions for this preconditioner, increasing the rank of the preconditioner substantially reduces the number of CG iterations required to achieve convergence.

In the bottom of \autoref{fig:precond_results}, we confirm that these more accurate solves indeed have an effect on the final test MAE.
We plot, as a function of the total wallclock time required to compute predictions, the test MAE resulting from using no preconditioner and from using a rank 5 preconditioner.
The wallclock time is varied by changing the number of CG iterations used to compute the predictive mean.
We observe that, because such a low rank preconditioner is sufficient, using preconditioning results in significantly more accurate solves while having virtually no impact on the running time of each CG iteration.
Consequentially, we recommend always using the pivoted Cholesky preconditioner with BBMM since it has virtually no wall-clock overhead and rapidly accelerates convergence.

%!TEX root=main_arxiv.tex
\section{Discussion}
In this paper, we discuss a novel framework for Gaussian process inference (\mmacro{}) based on blackbox matrix-matrix multiplication routines with kernel matrices.
We have implemented this framework and several state-of-the-art GP models in our new publicly available {\href{https://gpytorch.ai}{GPyTorch}} package.

\paragraph{Non-Gaussian likelihoods.}
Although this paper primarily focuses on the regression setting, \mmacro{} is fully compatible with variational techniques such as \cite{hensman2015scalable,wilson2016stochastic}, which are also supported in GPyTorch.
These approaches require computing the variational lower bound (or ELBO) rather than the GP marginal log likelihood \eqref{eq:log_lik_and_deriv}. We leave the exact details of the ELBO derivation to other papers (e.g. \cite{hensman2015scalable}).
However, we note that a single call to \mcgacro{} can be used to compute the KL divergence between two multivariate Gaussians, which is the most computationally intensive term of the ELBO.

\paragraph{Avoiding the Cholesky decomposition.}
A surprising and important take-away of this paper is that it is beneficial to avoid the Cholesky decomposition for GP inference, even in the exact GP setting.
The basic algorithm for the Cholesky decomposition (described in \autoref{app:pivoted_cholesky}) involves a divide-and conquer approach that can prove ill-suited for parallel hardware.
Additionally, the Cholesky decomposition performs a large amount of computation to get a linear solve when fast approximate methods suffice.
Ultimately, the Cholesky decomposition of a full matrix takes $\bigo{n^3}$ time while CG takes $\bigo{n^2}$ time.
Indeed, as shown in \autoref{fig:cg_error}, CG may even provide \emph{better} linear solves than the Cholesky decomposition.
%We liken this difference to the practical speed differences observed between second order methods like Newton's method and SGD in other settings.
While we use a pivoted version of this algorithm for preconditioning, we only compute the first five rows of this decomposition.
By terminating the algorithm very early, we avoid the computational bottleneck and many of the numerical instabilities.

It is our hope that this work dramatically reduces the complexity of implementing new Gaussian process models,
while allowing for inference to be performed as efficiently as possible.

\section*{Acknowledgements}
JRG and AGW are supported by NSF IIS-1563887 and by Facebook Research.
GP and KQW are supported in part by the III-1618134, III-1526012,
IIS-1149882, IIS-1724282, and TRIPODS-1740822 grants from the National Science
Foundation.
In addition, they are supported by the Bill and Melinda Gates Foundation, the Office of Naval Research, and SAP America Inc.

{\small
  \bibliographystyle{abbrvnat}
  \bibliography{citations}
}

%%%%%%%%%%%%%%%%%%%%%%%%%%%%%%%%%%%%%%%%%%%
%
% Supplementary Information
%
%%%%%%%%%%%%%%%%%%%%%%%%%%%%%%%%%%%%%%%%%%%

\clearpage

% Supplementary Information hacks from http://jshodges.com/index.php?qs=kb_001

\makeatletter
  % Tables and figures
  \setcounter{table}{0}
  \renewcommand{\thetable}{S\arabic{table}}%
  \setcounter{figure}{0}
  \renewcommand{\thefigure}{S\arabic{figure}}%
  % Equations
  \setcounter{equation}{0}
  \renewcommand\theequation{S\arabic{equation}}
  % Citations
  \renewcommand{\bibnumfmt}[1]{[S#1]}
  % Pages
  %\setcounter{page}{1}

  % Renew title
  \newcommand{\suptitle}{Supplementary Information for: \titl}
  \renewcommand{\@title}{\suptitle}
  \newcommand{\thanks}[1]{\footnotemark[1]}
  \renewcommand{\@author}{\authorinfo}

  % Make title
  \par
  \begingroup
    \renewcommand{\thefootnote}{\fnsymbol{footnote}}
    \renewcommand{\@makefnmark}{\hbox to \z@{$^{\@thefnmark}$\hss}}
    \renewcommand{\@makefntext}[1]{%
      \parindent 1em\noindent
      \hbox to 1.8em{\hss $\m@th ^{\@thefnmark}$}#1
    }
    \thispagestyle{empty}
    \@maketitle
    \@thanks
  \endgroup
  \let\maketitle\relax
  \let\thanks\relax
\makeatother

\appendix

%!TEX root=../main.tex
\section{Analysis of the modified CG algorithm, \mcgacro{}}
\label{app:mod_cg}

\paragraph{Linear conjugate gradients} is a widely popular algorithm from numerical linear algebra \cite{demmel1997applied,saad2003iterative,golub2012matrix} for rapidly performing matrix solves $\A^{-1} \bb$.
One view of CG is that it obtains matrix solves by solving a quadratic optimization problem:
\begin{equation}
  \A^{-1} \bb = \argmin_{\bu} f( \bu ) = \argmin_{\bu} \left( \frac{1}{2} \bu^\top \A \bu - \bu^\top \bb \right).
  \label{eq:cg_objective}
\end{equation}
CG iteratively finds the solution to \eqref{eq:cg_objective}.
After $n$ iterations, CG is guaranteed in exact arithmetic to find the exact solution to the linear system.
However, each iteration of conjugate gradients produces an approximate solve $\bu_k$ that can be computed with a simple recurrence.
We outline this recurrence in \autoref{alg:std_pcg}.\footnote{
  Note that this algorithm assumes a preconditioner.
  In absence of a preconditioner, one can set $\P^{-1} = \eye$.
}
\begin{algorithm2e}[H]
  \SetAlgoLined
  \SetKwInOut{Input}{Input}
  \SetKwInOut{Output}{Output}
  \newlength\inputlen
  \newcommand\NextInput[1]{%
    \settowidth\inputlen{\Input{}}%
    \setlength\hangindent{1.5\inputlen}%
    \hspace*{\inputlen}#1\\
  }
  \newcommand\graycomment[1]{\footnotesize\ttfamily\textcolor{gray}{#1}}
  \SetCommentSty{graycomment}
  \SetKw{Break}{break}
  \SetKwData{tol}{tolerance}
  \SetKwFunction{mvmkxx}{mvm\_$\A$}
  \SetKwFunction{mvmprec}{$\P^{-1}$}
  \SetKwFunction{size}{size}
  \caption{Standard preconditioned conjugate gradients (PCG).}
  \label{alg:std_pcg}
    \Input{\mvmkxx{} -- function for matrix-vector multiplication (MVM) with matrix $\A$}
    \NextInput{$\bb$ -- vector to solve against}
    \NextInput{\mvmprec{} -- function for preconditioner}
    \Output{$\A^{-1} \bb$.}
    \BlankLine
	  $\bu_0$ $\gets$ $\mathbf 0$
    \tcp{Current solution}
    $\br_0$ $\gets$ \mvmkxx{$\bu_0$} - \bb
    \tcp{Current error}
    $\bz_0$ $\gets$ \mvmprec{$\br_0$}
    \tcp{Preconditioned error}
    $\bd_0$ $\gets$ {$\bz_0$}
    \tcp{``Search'' direction for next solution}
    \BlankLine
    \For{$j \gets 0$ \KwTo T}{
      $\bv_{j}$ $\gets$ \mvmkxx{ $\bd_{j-1}$ }
      \\
      $\alpha_j$ $\gets$ $( \br_{j-1}^\top \bz_{j-1} ) / ( \bd_{j-1}^\top \bv_{j} )$
      \\
      $\bu_j$ $\gets$ $\bu_{j-1} + \alpha_{j} \bd_{j-1}$
      \\
      $\br_j$ $\gets$ $\br_{j-1} - \alpha_{j} \bv_{j}$
      \\
      \lIf{$\left\Vert \br_{j} \right\Vert_2$ $<$ \tol}{
        \Return{$\bu_j$}
      }
      $\bz_{j}$ $\gets$ \mvmprec{ $\br_{j}$ }
      \\
      $\beta_j$ $\gets$ $( \bz_{j}^\top \bz_{j} ) / ( \bz_{j-1}^\top \bz_{j-1} )$
      \\
      $\bd_{j}$ $\gets$ $\bz_{j} - \beta_j \bd_{j-1}$
    }
    \BlankLine
    \Return{$\bu_{j+1}$}
\end{algorithm2e}
Conjugate gradients avoids explicitly computing the matrix $A$, and uses only a matrix vector multiply routine instead. This leads to the following observation about CG's running time:
\begin{observation}[Runtime and space of conjugate gradients]
  When performing Conjugate gradients to solve $\A^{-1}\bb$, the matrix $\A$ is only accessed through \emph{matrix-vector multiplications} (MVMs) with $\A$.
  The time complexity of $p$ iterations of CG is therefore $\bigo{p \: \mvm{\A}}$, where $\mvm{\A}$ is the cost of one MVM with $\A$.
  The space complexity is $\bigo{n}$ (assuming $\A \in \reals^{n \times n})$.
\end{observation}
This is especially advantageous is $\A$ is a sparse or structured matrix with a fast MVM algorithm.
Additionally, CG has remarkable convergence properties, and in practice returns solves accurate to nearly machine precision in $p \ll n$ iterations.
The error of the approximation between the $p$th iterate and the optimal solution $\bu^{*}$, $\Vert \bu_{k} - \bu^{*}\Vert$, depends more on the conditioning of the matrix than on the size of the matrix $\A$.
Formally, the error can be bounded in terms of an \emph{exponential decay} involving the \emph{condition number} $\kappa(\A)=\Vert \A \Vert_{2} \Vert \A^{-1} \Vert_{2}$:
\begin{observation}[Convergence of conjugate gradients \cite{demmel1997applied,golub2012matrix,saad2003iterative}]
  \label{obs:cg_convergence}
  Let $\A$ be a positive definite matrix, and let $\bu^{*}$ be the optimal solution to the linear system $\A^{-1}\bb$.
  The error of the $p$th iterate of conjugate gradients $\bu_{p}$ can be bounded as follows:
  \begin{equation}
    \begin{aligned}
    \Vert \bu^{*} - \bu_{p} \Vert_{\A}
    &\leq 2 \left( \left( {\sqrt{\kappa\left(\A\right)} - 1} \right) / \left( {\sqrt{\kappa\left(\A\right)} + 1} \right) \right)^{p}\Vert \bu^{*} - \bu_{0} \Vert_{\A},
  \end{aligned}
  \end{equation}
\end{observation}
where $\Vert \cdot \Vert_{\A}$ is the $A$ norm of a vector -- i.e. $\Vert \bv \Vert_{\A} = ( \bv^\top \: A \bv )^{1/2}$ \cite{demmel1997applied,golub2012matrix,saad2003iterative}.

\paragraph{Modified Batched Conjugate Gradients (\mcgacro{}),} which we introduce in \autoref{sec:method}, makes two changes to standard CG.
In particular, it performs multiple solves $\A^{-1} \B = [\A^{-1} \bb_1, \ldots, \A^{-1} \bb_t]$ simultaneously using {\bf matrix-matrix multiplication} (MMM), and it also returns Lanczos tridiagonalization matrices associated with each of the solves.
The Lanczos tridiagonalization matrices are used for estimating the log determinant of $\A$.
\mcgacro{} is outlined in \autoref{alg:mod_pcg}:
{\ }

\boldmatrixtrue

\newcommand{\colornew}{red}
\newcommand{\colormat}{blue}

\begin{algorithm2e}[H]
  \SetAlgoLined
  \SetKwInOut{Input}{Input}
  \SetKwInOut{Output}{Output}
  \newcommand\NextInput[1]{%
    \settowidth\inputlen{\Input{}}%
    \setlength\hangindent{1.5\inputlen}%
    \hspace*{\inputlen}#1\\
  }
  \newcommand\graycomment[1]{\footnotesize\ttfamily\textcolor{gray}{#1}}
  \SetCommentSty{graycomment}
  \SetKw{Break}{break}
  \SetKwData{tol}{tolerance}
  \SetKwFunction{mmmkxx}{mmm\_$\A$}
  \SetKwFunction{mmmprec}{$\trainP^{-1}$}
  \SetKwFunction{size}{size}
  \SetKwFunction{diag}{diag}
  \caption{Modified preconditioned conjugate gradients (PCG).}
  \label{alg:mod_pcg}
    \Input{{\color{\colormat} \mmmkxx{}} -- function for {\color{\colormat} matrix-matrix multiplication} with $\A$}
    \NextInput{{\color{\colormat} $\B$ -- $n \times t$ matrix} to solve against}
    \NextInput{\mmmprec{} -- func. for preconditioner}
    \Output{{\color{\colormat} $\A^{-1} \B$}, {\color{\colornew} $\tilde \T_{1}$, $\ldots$, $\tilde \T_{t}$}.}
    \BlankLine
    {\color{\colormat} $\U_0$ $\gets$ $\mathbf 0$}
    \tcp{Current solutions}
    {\color{\colormat} $\R_0$ $\gets$ \mmmkxx{$\U_0$} - \B}
    \tcp{Current errors}
    {\color{\colormat} $\Z_0$ $\gets$ \mvmprec{$\R_0$}}
    \tcp{Preconditioned errors}
    {\color{\colormat} $\D_0$ $\gets$ {$\Z_0$}}
    \tcp{``Search'' directions for next solutions}
    {\color{\colornew} $\tilde \T_{1}, \ldots \tilde \T_{t}$ $\gets$ $0$}
    \tcp{Tridiag matrices}
    \vspace{0.32em}
    \For{$j \gets 0$ \KwTo t}{
      {\color{\colormat} $\V_{j}$ $\gets$ \mmmkxx{ $\D_{j-1}$ }}
      \\
      {\color{\colormat} $\vec \alpha_j$ $\gets$ ${( \R_{j-1} \circ \Z_{j-1} )^\top \mathbf 1}/{( \D_{j-1} \circ \V_{j} )^\top \mathbf 1}$}
      \\
      {\color{\colormat} $\U_j$ $\gets$ $\U_{j-1} +$ \diag{$\vec \alpha_{j}$} $\D_{j-1}$}
      \\
      {\color{\colormat} $\R_j$ $\gets$ $\R_{j-1} -$ \diag{$\vec \alpha_{j}$} $\V_{j}$}
      \\
      {\color{\colormat} \lIf{$\forall i$ \: $\left\Vert \br^{(i)}_{j} \right\Vert_2$ $<$ \tol}{
        \Return{$\U_j$}
      }}
      {\color{\colormat} $\Z_{j}$ $\gets$ \mmmprec{ $\R_{j}$ }}
      \\
      {\color{\colormat} $\vec \beta_j$ $\gets$ ${( \Z_{j} \circ \Z_{j} )^\top \mathbf 1}/{( \Z_{j-1} \circ \Z_{j-1} )^\top \mathbf 1}$}
      \\
      {\color{\colormat} $\D_{j}$ $\gets$ $\Z_{j} -$ \diag{$\vec \beta_j$} $\D_{j-1}$}
      \\
      {\color{\colornew} $\forall i$ \: $\tilde [\T_{i}]_{j,j}$ $\gets$ $1/[\vec \alpha_{j}]_{i} + [\vec \beta_{j-1}]_{i}/[\vec \alpha_{j-1}]_{i}$}
      \\
      {\color{\colornew} $\forall i$ \: $\tilde [\T_{i}]_{j-1,j}$, $\tilde [\T_{i}]_{j,j-1}$ $\gets$ $\sqrt{[\vec \beta_{j-1}]_{i}}/[\vec \alpha_{j}]_{i}$}
    }
    \Return{{\color{\colormat}$\U_{j+1}$}, {\color{\colornew} $\tilde \T_{1}, \ldots \tilde \T_{t}$ $\gets$ $0$}}
\end{algorithm2e}

\boldmatrixfalse
Note that {\color{\colormat} \colormat} represents an operation that is converted from a vector operation to a matrix operation.
{\color{\colornew} \colornew} is an addition to the CG algorithm to compute the tridiagonalization matrices.

In this section, we will derive the correctness of the modified batched CG algorithm.
We will show that the matrix operations perform multiple solves in parallel.
Additionally, we will show that the tridiagonal matrices $\tilde \T_1, \ldots, \tilde \T_t$ correspond to the tridiagonal matrices of the Lanczos algorithm \cite{lanczos1950iteration}.

\subsection{Adaptation to multiple right hand sides.}
The majority of the lines in Algorithm 2 are direct adaptations of lines from Algorithm 1 to handle multiple vectors simultaneously.
We denote these lines in {\color{\colormat} \colormat}.
For example, performing
\begin{equation*}
\V_{j} \gets \mmmkxx{ $P_{j-1}$ }
\end{equation*}
is equivalent to performing $\bv_{j} \gets \mvmkxx{ $\bd_{j-1}$ }$ for each column of $P_{j-1}$.
Thus we can replace multiple MVM calls with a single MMM call.

In standard CG, there are two scalar coefficient used during each iteration: $\alpha_j$ and $\beta_j$ (see \autoref{alg:std_pcg}).
In \mcgacro{}, each solve $\bu_1, \ldots, \bu_t$ uses different scalar values.
We therefore now have \emph{two coefficient vectors}: $\vec \alpha_j \in \reals^t$ and $\vec \beta_j \in \reals^t$, where each of the entries corresponds to a single solve.
There are two types of operations involving these coefficients:
\begin{enumerate}
  \item Updates (e.g. {\color{\colormat} $\vec \alpha_j$ $\gets$ ${( \R_{j-1} \circ \Z_{j-1} )^\top \mathbf 1}/{( \D_{j-1} \circ \V_{j} )^\top \mathbf 1}$})
  \item Scalaing (e.g. {\color{\colormat} $\U_j$ $\gets$ $\U_{j-1} +$ \diag{$\vec \alpha_{j}$} $\D_{j-1}$})
\end{enumerate}
The update rules are batched versions of the update rules in the standard CG algorithm.
For example:
\begin{equation*}
  \left[ \begin{array}{c}
    \left[ \vec \alpha_j \right]_1
    \\
    \vdots
    \\
    \left[ \vec \alpha_j \right]_t
  \end{array} \right]
  = \frac{( \R_{j-1} \circ \Z_{j-1} )^\top \mathbf 1}{( \D_{j-1} \circ \V_{j} )^\top \mathbf 1}
  = \left[ \begin{array}{c}
        \frac{\left( [\R_{j-1}]_{1} \circ [\Z_{j-1}]_{1} \right) \mathbf 1}
        {\left( [\D_{j-1}]_{1} \circ [\V_{j}]_{1} \right) \mathbf 1}
        \\
        \vdots
        \\
        \frac{\left( [\R_{j-1}]_{t} \circ [\Z_{j-1}]_{t} \right) \mathbf 1}
        {\left( [\D_{j-1}]_{t} \circ [\V_{j}]_{t} \right) \mathbf 1}
     \end{array} \right]
  = \left[ \begin{array}{c}
        \frac{[\R_{j-1}]_{1}^\top [\Z_{j-1}]_{1}}
        {[\D_{j-1}]_{1}^\top [\V_{j}]_{1}}
        \\
        \vdots
        \\
        \frac{[\R_{j-1}]_{t}^\top [\Z_{j-1}]_{t}}
        {[\D_{j-1}]_{t}^\top [\V_{j}]_{t}}
     \end{array} \right],
\end{equation*}
using the identity $(\bv \cdots \bv') \mathbf 1 = \bv^\top \bv'$.
Thus these updates are batched versions of their non-batched counterparts in \autoref{alg:std_pcg}.
Similarly, for scaling,
\begin{align*}
  \left[ \begin{array}{ccc}
    \left[\U_j \right]_1 & \cdots & \left[ \U_j \right]_t
  \end{array} \right]
  &=
  \U_j = \U_{j-1} + \text{diag}(\alpha_j) \D_{j-1}
  \\
  &=
  \left[ \begin{array}{ccc}
    \left[\U_{j-1} \right]_1 & \cdots & \left[ \U_{j-1} \right]_t
  \end{array} \right]
  +
  \left[ \begin{array}{ccc}
    \left[ \alpha_{j} \right]_1 \left[\D_{j-1} \right]_1 & \cdots & \left[ \alpha_{j} \right]_t \left[ \D_{j-1} \right]_t
  \end{array} \right].
\end{align*}
This these scaling operations are also batched versions of their counterparts in \autoref{alg:std_pcg}.
\mcgacro{} is therefore able to perform all solve operations in batch, allowing it to perform multiple solves at once.

\subsection{Obtaining Lanczos tridiagonal matrices from \mcgacro{}.}

To motivate the Lanczos tridiagonal matrices $\tilde \T_1, \ldots, \tilde \T_t$ from \mcgacro{}, we will first discuss the Lanczos algorithm.
Then, we will discuss how \mcgacro{} recovers these matrices.

\paragraph{The Lanczos algorithm} \cite{lanczos1950iteration} is an iterative MVM-based procedure to obtain a \emph{tridiagonalization} of a symmetric matrix $\A$.
A tridiagonalization is a decomposition $\A = \Q \T \Q^{\top}$ with $\Q \in \reals^{n \times n}$ orthonormal and $\T \in \reals^{n \times n}$ tridiagonal -- i.e.
\begin{equation}
    T = \left[\begin{array}{ccccc}
    d_{1} & s_{1} &  & & 0 \\
    s_{1} & d_{2} & s_{2} &  &  \\
     & s_{2} & d_{3} & \ddots & \\
     &       & \ddots & \ddots & s_{n-1} \\
     0 &       &        & s_{n-1} & d_{n}

  \end{array}\right].
\end{equation}
The exact $\Q$ and $\T$ matrices are uniquely determined by a \emph{probe vector} $\bz$ -- which is the first column of $\Q$.
The Lanczos algorithm iteratively builds the rest of $\Q$ and $\T$ by forming basis vectors in the \emph{Krylov subspace} -- i.e.
\begin{equation}
  \text{span} \left\{ \bz, \A \bz, \A^2 \bz, \ldots, \A^{n-1} \bz \right\},
  \label{eq:lanczos}
\end{equation}
and applying Gram-Schmidt orthogonalization to these basis vectors.
The orthogonalized vectors are collected in $\Q$ and the Gram-Schmidt coefficients are collected in $\T$.
\citet{lanczos1950iteration} shows that $n$ iterations of this procedure produces an exact tridiagonalization $\A = \Q \T \Q^\top$.
$p$ iterations yields a low-rank \emph{approximate tridiagonalization} $\A \approx \tilde \Q \tilde \T \tilde \Q^\top$,
where $\tilde \Q \in \reals^{n \times p}$ and $\tilde \T \in \reals^{p \times p}$.

\paragraph{Connection between the Lanczos algorithm and conjugate gradients.}

There is a well-established connection between the Lanczos algorithm and conjugate gradients \cite{demmel1997applied,golub2012matrix,saad2003iterative}.
In fact, the conjugate gradients algorithm can even be \emph{derived} as a byproduct of the Lanczos algorithm.
\citet{saad2003iterative} and others show that it is possible to recover the $\tilde T$ tridigonal Lanczos matrix by \emph{reusing coefficients} generated in CG iterations.
In particular, we will store the $\alpha_j$ and $\beta_j$ coefficients from \autoref{alg:std_pcg}.
\begin{observation}[Recovering Lanczos tridiagonl matrices from standard CG \cite{saad2003iterative}]
  Assume we use $p$ iterations of standard preconditioned conjugate gradients to solve $A^{-1} \bz$ with preconditioner $P$.
  Let $\alpha_1, \ldots, \alpha_p$ and $\beta_1, \ldots, \beta_p$ be the scalar coefficients from each iteration (defined in \autoref{alg:std_pcg}).
  The matrix
  \begin{equation}
    \left[\begin{array}{ccccc}
      \frac{1}{\alpha_1} & \frac{\sqrt{\beta_1}}{\alpha_1} &  & & 0 \\
      \frac{\sqrt{\beta_1}}{\alpha_1} & \frac{1}{\alpha_2} + \frac{\beta_1}{\alpha_1} & \frac{\sqrt{\beta_2}}{\alpha_2} &  &  \\
      & \frac{\sqrt{\beta_2}}{\alpha_2} & \frac{1}{\alpha_3} + \frac{\beta_2}{\alpha_2} & \frac{\sqrt{\beta_3}}{\alpha_3} &  \\
      &       & \ddots & \ddots & \frac{\sqrt{\beta_{m-1}}}{\alpha_{m-1}} \\
      0 &       &        & \frac{\sqrt{\beta_{m-1}}}{\alpha_{m-1}} & \frac{1}{\alpha_m} + \frac{\beta_{m-1}}{\alpha_{m-1}}
    \end{array}\right]
  \end{equation}
  is equal to the Lanczos tridiagonal matrix $\tilde \T$, formed by running $p$ iterations of Lanczos to achieve $\tilde\Q^{\top}  \P^{-1} \A \tilde \Q= \tilde \T$) with probe vector $\bz$.
\end{observation}
(See \cite{saad2003iterative}, Section 6.7.3.)
In other words, we can recover the Lanczos tridiagonal matrix $\tilde T$ simply by running CG.
Our \mcgacro{} algorithm simply exploits this fact.
The final two lines in {\color{\colornew} \colornew} in \autoref{alg:mod_pcg} use the $\vec \alpha_j$ and $\vec \beta_j$ coefficients to form $t$ tridiagonal matrices.
If we are solving the systems $\A^{-1}[\bb_1, \ldots, \bb_t]$, then the resulting tridiagonal matrices correspond to the Lanczos matrices with probe vectors $\bb_1, \ldots, \bb_t$.

\section{Runtime analysis of computing inference terms with \mcgacro{}}
\label{app:method_runtime}

We first briefly analyze the running time of \mcgacro{} (\autoref{alg:mod_pcg}) itself.
The algorithm performs matrix multiplies with $\trainK$ once before the loop and once during every iteration of the loop.
Therefore, the running time of \mcgacro{} is at least $\bigo{p\mmm{\trainK}}$, where $\mmm{\trainK}$ is the time to multiply $\trainK$ by an $n \times t$ matrix.

For the remainder of the algorithm, all matrices involved ($U_{j},V_j,R_j,Z_j,P_j$) are $n \times t$ matrices.
All of the lines involving only these matrices perform operations that require $\bigo{nt}$ time.
For example, elementwise multiplying $Z_{j} \circ Z_{j}$ accesses each element in $Z_{j}$ once, and and then multiplying it by the vector of ones similarly accesses every element in the matrix once.
Multiplying $V_{j}$ by the diagonal matrix with $\ba_{j}$ on the diagonal takes $\bigo{nt}$ time, because we multiply every element $[V_{j}]_{ik}$ by $[\ba_{j}]_{i}$.
Therefore, all other lines in the algorithm are dominated by the matrix multiply with $\trainK$, and the total running time is also $\bigo{p\mmm{\trainK}}$.
Furthermore, because these intermediate matrices are $n \times t$, the space requirement (beyond what is required to store $\trainK$) is also $\bigo{nt}$.

We will now show that, after using \mcgacro{} to produce the solves and tridiagonal matrices, recovering the three inference terms takes little additional time and space.
To recap, we run \mcgacro{} to recover
\begin{equation*}
  \left[\begin{array}{cccc}\bu_{0} & \bu_{1} & \cdots & \bu_{t}\end{array}\right] = \trainK^{-1}\left[\begin{array}{cccc}\by & \bz_{1} & \cdots & \bz_{t}\end{array}\right] \:\:\:\:\: \text{and} \:\:\:\:\: \tilde{\T}_{1},...,\tilde{\T}_{t}.
\end{equation*}
where $\bz_i$ are random vectors and $\tilde \T_i$ are their associated Lanczos tridiagonal matrices.

\paragraph{Time complexity of $\trainK^{-1}\y$.}
This requires no additional work over running \mcgacro{}, because it is the first output of the algorithm.

\paragraph{Time complexity of $\tr{\trainK^{-1}\frac{d\trainK}{d\theta}}$.}
\mcgacro{} gives us access to $\trainK^{-1}[\bz_{1} \: \ldots \: \bz_{t}]$.
Recall that we compute this trace as:
\begin{equation}
  \tr{\trainK^{-1}\frac{dK}{d\theta}} \approx \frac{1}{t}\sum_{i=1}^{t}(\bz_{i}\trainK^{-1})\left(\frac{d\trainK}{d\theta}\bz_{i}\right)
\end{equation}
We can get $\frac{d\trainK}{d\theta}\bz_{i}$ by performing a single matrix multiply $\frac{d\trainK}{d\theta}[\bz_{1} \: \ldots \: \bz_{t}]$, requiring $\mmm{\frac{d\trainK}{d\theta}}$.
(We assume that $\mmm{\frac{d\trainK}{d\theta}} \approx \mmm{\trainK}$, which is true for exact GPs and all sparse GP approximations.)
After this, we need to perform $t$ inner products between the columns of this result and the columns of $\trainK^{-1}[\bz_{1} \: \ldots \: \bz_{t}]$, requiring $\bigo{tn}$ additional time.
Therefore, the running time is still dominated by the running time of \mcgacro{}.
The additional space complexity involves the $2t$ length $n$ vectors involved in the inner products, which is negligible.

\paragraph{Time complexity of $\log \vert \trainK \vert$.}
\mcgacro{} gives us $p \times p$ tridiagonal matrices  $\tilde{T}_{1},...,\tilde{T}_{t}$. To compute the log determinant estimate, we must compute $e_{1}^{\top}\log \tilde{T}_{i} e_{1}$ for each $i$. To do this, we eigendecompose $\tilde{T}_{i}=V_{i}\Lambda_{i}V_{i}^{\top}$, which can be done in $\bigo{p^{2}}$ time for tridiagonal matrices, and compute
\begin{equation}
  e_1^\top V_i\log \Lambda_i V_i^\top e_1
\end{equation}
where now the $\log$ is elementwise over the eigenvalues. Computing $V_i^\top e_1$ simply gets the first row of $V_i$, and $\log \Lambda$ is diagonal, so this requires only $\bigo{p}$ additional work.

The total running time post-\mcgacro{} is therefore dominated by the $\bigo{tp^{2}}$ time required to eigendecompose each matrix. This is again significantly lower than the running time complexity of \mcgacro{} itself. The space complexity involves storing $2t$ $p \times p$ matrices (the eigenvectors), or $\bigo{tp^{2}}$.

%!TEX root=../main.tex
\section{The Pivoted Cholesky Decomposition}
\label{app:pivoted_cholesky}

In this section, we review a full derivation of the \emph{pivoted Cholesky decomposition} as used for precondtioning in our paper.
To begin, observe that the standard Cholesky decomposition can be seen as producing a sequentially more accurate low rank approximation to the input matrix $K$.
In particular, the Cholesky decomposition algorithm seeks to decompose a matrix $K$ as:
\begin{equation}
 \label{eq:cholesky_full}
 \left[\begin{array}{cc}K_{11} & K_{12} \\ K_{12}^{\top} & K_{22}\end{array}\right] = \left[\begin{array}{cc}L_{11} & 0 \\ L_{21} & L_{22}\end{array}\right] \left[\begin{array}{cc}L_{11}^{\top} & L_{21}^{\top} \\ 0 & L_{22}^{\top}\end{array}\right]
\end{equation}
Note that that $K_{11} = L_{11}L_{11}^{\top}$, $K_{12} = L_{11}L_{21}^{\top}$, and $K_{22} = L_{21}L_{21}^{\top} + L_{22}L_{22}^{\top}$.
From these equations, we can obtain $L_{11}$ by recursively Cholesky decomposing $K_{11}$, $L_{21}^{\top}$ by computing $L_{21}^{\top} = L_{11}^{-1}K_{12}$, and finally $L_{22}$ by Cholesky decomposing the Schur complement $S = K_{22} - L_{21}L_{21}^{\top}$.

Rather than compute the full Cholesky decomposition, we can view each iteration of the Cholesky decomposition as producing a slightly higher rank approximation to the matrix $K$.
In particular, if $K = \left[\begin{array}{cc}k_{11} & \bb^{\top} \\ \bb & K_{22}\end{array}\right]$, then $L_{11} = \sqrt{k_{11}}$, $L_{21}=\frac{1}{\sqrt{k_{11}}}\bb$, and the Schur complement is $S = K_{22} - \frac{1}{k_{11}}\bb\bb^{\top}$.
Therefore:
\begin{align}
  K &= \frac{1}{k_{11}}\left[\begin{array}{c}k_{11} \\ \bb\end{array}\right]\left[\begin{array}{c}k_{11} \\ \bb\end{array}\right]^{\top} + \left[\begin{array}{cc}0 & 0 \\ 0 & S\end{array}\right] \\
    &= \bq_{1}\bq_{1}^{\top} + \left[\begin{array}{cc}0 & 0 \\ 0 & S\end{array}\right] \label{eq:chol_incomplete}.
\end{align}
Because the Schur complement is positive definite \cite{harbrecht2012low}, we can continue by recursing on the $n-1 \times n - 1$ Schur complement $S$ to get another vector.
In particular, if $S = \bq_{2}\bq_{2}^{\top} + \left[\begin{array}{cc}0 & 0 \\ 0 & S'\end{array}\right]$, then:
\begin{align}
  K  = \bq_{1}\bq_{1}^{\top} + \left[\begin{array}{c}0 \\ \bq_{2}\end{array}\right]\left[\begin{array}{c}0 \\ \bq_{2}\end{array}\right]^{\top} + \left[\begin{array}{cc}0 & 0 \\ 0 & S'\end{array}\right]
\end{align}
In general, after $k$ iterations, defining $\hat{\bq}_{i} = \left[\begin{array}{c}\mathbf{0} \\ \bq_{i}\end{array}\right]$ from this procedure, we obtain
\begin{align}
  K  = \sum_{i=1}^{k}\hat{\bq}_{i}\hat{\bq}_{i}^{\top} + \left[\begin{array}{cc}0 & 0 \\ 0 & S_{k}\end{array}\right].
\end{align}
The matrix $P_{k}=\sum_{i=1}^{k}\hat{\bq}_{i}\hat{\bq}_{i}^{\top}$ can be viewed as a low rank approximation to $K$, with
$$
  \Vert K - P_{k} \Vert_{2} = \left\Vert \left[\begin{array}{cc}0 & 0 \\ 0 & S_{k}\end{array}\right] \right\Vert_{2}.
$$

To improve the accuracy of the low rank approximation, one natural goal is to minimize the norm of the Schur complement, $\Vert S_{i} \Vert$, at each iteration.
\citet{harbrecht2012low} suggest to permute the rows and columns of $S_{i}$ (with $S_{0} = K$) so that the upper-leftmost entry in $S_{i}$ is the maximum diagonal element.
In the first step, this amounts to replacing $K$ with $\pi_{1}K\pi_{1}$, where $\pi_{1}$ is a permutation matrix that swaps the first row and column with whichever row and column corresponds to the maximum diagonal element of $K$.
Thus:
\begin{align}
  \pi_{1}K\pi_{1} = \bq_{1}\bq_{1}^{\top} + \left[\begin{array}{cc}0 & 0 \\ 0 & S\end{array}\right].
\end{align}
To proceed, one can apply the same pivoting rule to $S$ to achieve $\pi_{2}$.
Defining $\hat{\pi}_{2} = \left[\begin{array}{cc}1 & 0 \\ 0 & \pi_{2}\end{array}\right]$, then we have:
\begin{align}
  \hat{\pi}_{2}\pi_{1}K\pi_{1}\hat{\pi}_{2} = \hat{\pi}_{2}\bq_{1}\bq_{1}^{\top}\hat{\pi}_{2} + \hat{\bq}_{2}\hat{\bq}_{2}^{\top} + \left[\begin{array}{cc}0 & 0 \\ 0 & S_{2}\end{array}\right].
\end{align}
To obtain a rank two approximation to $K$ from this, we multiply from the left and right by all permutation matrices involved:
\begin{align}
  K = \pi_{1}\bq_{1}\bq_{1}^{\top}\pi_{1} + \pi_{1}\hat{\pi}_{2}\hat{\bq}_{2}\hat{\bq}_{2}^{\top}\hat{\pi}_{2}\pi_{1} + E_{2}.
\end{align}
In general, after $k$ steps, we obtain:
\begin{align}
  K = \sum_{i=1}^{k}(\mathbb{Q}_{i}\hat{\bq}_{i})(\mathbb{Q}_{i}\hat{\bq}_{i})^{\top} + E_{k},
\end{align}
where $\mathbb{Q}_{i} = \prod_{j=1}^{i}\hat{\pi}_{j}$.
By collecting these vectors in to a matrix, we have that $K = L_{k}L_{k}^{\top} + E_{k}$, and thus $K \approx L_{k}L_{k}^{\top}$.

\subsection{Running time of the pivoted Cholesky decomposition.}
Let $L_k L_k^\top$ be the rank $k$ pivoted Cholesky decomposition of $K_{XX}$.
We now analyze the time complexity of computing the pivoted Cholesky decomposition and using it for preconditioning.
We will prove the claims made about the pivoted Cholesky decomposition made in the main text which are restated here:
\begin{observation}[Properties of the Pivoted Cholesky decomposition]
  {\ }
  \begin{enumerate}
    \item Let $L_k L_k^\top$ be the rank $k$ pivoted Cholesky decomposition of $K_{XX}$.
      It can be computed in $\bigo{\row{K_{XX}}k^{2}}$ time, where $\row{K_{XX}}$ is the time required to retrieve a single row of $K_{XX}$.
    \item Linear solves with $\trainP = L_{k}L_{k}^{\top} + \sigma^{2}I$ can be performed in $\bigo{nk^{2}}$ time.
    \item The log determinant of $\trainP$ can be computed in $\bigo{nk^{2}}$ time.
  \end{enumerate}
\end{observation}

\paragraph{Time complexity of computing $L_{k} L_k^\top$.}
In general, computing $L_k$ requires reading the diagonal of $K_{XX}$ and $k$ rows of the matrix.
For a standard positive definite matrix, \citet{harbrecht2012low} observes that this amounts to a $\bigo{nk^{2}}$ running time.
Given that the time requirement for an matrix-vector multiplication with a standard matrix $\trainK$ is $\bigo{n^2}$, computating the pivoted Cholesky decomposition is a negligible operation.

More generally if we wish to avoid computing the exact matrix $K_{XX}$, then the time requirement is $\bigo{\rho(A)k^2}$,
where $\rho(A)$ is the time required to access a row of $A$.
When applying the SoR approximation ($K_{XX} = K_{XU}K_{UU}^{-1}K_{XU}^{\top}$), we have that $\rho(K)=\bigo{nm}$.
Thus, for SGPR, the time complexity of computing the rank $k$ pivoted Cholesky decomposition is $\bigo{nmk^{2}}$ time.
Assuming that $k^2 \leq m$ or $k^2 \approx m$, this operation will cost roughly the same as a single MVM.

When applying the SKI approximation ($K_{XX}=WK_{UU}W^{\top}$), we have that $\rho(K_{XX})=\bigo{n}$.
In particular, the ith row of $K_{XX}$ is given by
$
   \left[K_{XX}\right]_{i} = \bw_{i}K_{UU}W^{\top}.
$
Rather than explicitly perform these multiplications using Toeplitz matrix arithmetic, we observe that $\bw_{i}K_{UU}$ is equivalent to summing four elements from each column of $K_{UU}$ (corresponding to the four non-zero elements of $\bw_i$).
Since elements of $K_{UU}$ can be accessed in $\bigo{1}$ time, this multiplication requires $\bigo{m}$ work.
After computing $\bv=\bw_{i}K_{UU}$, computing $\bv W^{\top}$ requires $\bigo{n}$ work due to the sparsity of $W^{\top}$.
Therefore, we can compute a pivoted Cholesky decomposition for a SKI kernel matrix in $\bigo{nk^{2}}$ time.
This time complexity is comparable to the MVM time complexity, which is also linear in $n$.

\paragraph{Time complexity of computing $\trainP_{k}^{-1}\by$ and $\log \vert \trainP_k \vert$.}
To compute solves with the preconditioner, we make use of the Woodbury formula.
Observing that $P_{k} = L_{k}L_{k}^{\top}$,
\begin{equation*}
  \trainP_{k}^{-1}\by = (L_{k}L_{k}^{\top} + \sigma^{2}I)^{-1}\by = \frac{1}{\sigma^{2}}\by - \frac{1}{\sigma^{4}}L_{k}(I - \frac{1}{\sigma^{2}}L_{k}^{\top}L_{k})^{-1}L_{k}^{\top}\by
\end{equation*}
Computing $L_{k}^{\top}\by$ takes $\bigo{nk}$ time. After computing the $k \times k$ matrix $I - \frac{1}{\sigma^{2}}L_{k}^{\top}L_{k}$ in $\bigo{nk^{2}}$ time, computing a linear solve with it takes $\bigo{k^{3}}$ time. Therefore, each solve with the preconditioner, $P_{k} + \sigma^{2}I$, requires $\bigo{nk^{2}}$ time total.
To compute the log determinant of the preconditioner, we make use of the matrix determinant lemma:
\begin{equation*}
  \log \vert \trainP_{k} \vert = \log \vert P_{k} + \sigma^{2}I \vert = \log \vert I - \frac{1}{\sigma^{2}}L_{k}^{\top}L_{k} \vert + 2 n \log \sigma
\end{equation*}
Since $I - \frac{1}{\sigma^{2}}L_{k}L_{k}^{\top}$ is a $k \times k$ matrix, we can compute the above log determinant in $\bigo{nk^{2}}$ time.

\paragraph{Time complexity of drawing samples from $\normaldist{0}{\trainP_k}$.}
We draw samples from $\normaldist{0}{\trainP_k}$ via the reparameterization trick \citep{kingma2014auto}.
If $\epsilon_1' \in \reals^k$ and $\epsilon_2' \in \reals^n$ are standard normal vectors,
then $\left( \L_k \epsilon_1' + \sigma \epsilon_2' \right)$ is a sample from $\normaldist{ 0 }{( \L_k \L_k^\top + \sigma^2 \I )}$.
Computing this requires a matrix-vector multiply with $\L_k$ for a total of $\bigo{nk}$ time.

%!TEX root=../main.tex
\section{Convergence Analysis of Pivoted Cholesky Preconditioned CG}
\label{app:theory}

In this section we prove \autoref{thm:cg_convergence_rbf}, which bounds the convergence of pivoted Cholesky-preconditioned CG for univariate RBF kernels.
\newtheorem*{thm:cg_convergence_rbf}{\autoref{thm:cg_convergence_rbf} (Restated)}
\begin{thm:cg_convergence_rbf}
  Let $K_{XX} \in \reals^{n \times n}$ be a $n \times n$ univariate RBF kernel, and let $L_k L_k^\top$ be its rank $k$ pivoted Cholesky decomposition.
  Assume we are using preconditioned CG to solve the system $\trainK^{-1} \y = (K_{XX} + \sigma^2 I)^{-1} \y$ with preconditioner $\trainP = (L_k L_k^\top + \sigma^2 I)$.
  Let $\bu_k$ be the $k^\textrm{th}$ solution of CG, and let $\bu^{*} = \trainK^{-1} \y$ be the exact solution.
  Then there exists some $b > 0$ such that:
  \begin{equation}
    \Vert \bu^{*} - \bu_{k} \Vert_{\trainK}
    \leq 2 \left(\frac{1}{1 + \bigo{n^{-1/2} \exp(kb/2)}}\right)^{p} \left\Vert \bu^{*} - \bu_{0} \right\Vert_{\trainK}.
  \end{equation}
\end{thm:cg_convergence_rbf}

\begin{proof}
Let $L_k L_k^\top$ be the rank $k$ pivoted Cholesky decomposition of a univariate RBF kernel $K_{XX}$.
  We begin by stating a well-known CG convergence result, which bounds error in terms of the \emph{conditioning number} $\kappa$ (see \autoref{obs:cg_convergence}):
\begin{equation}
  \label{eq:condition_number_convergence}
  \left\Vert \bu^* - \bu_{k} \right\Vert_{\trainK}
  \leq 2 \left( \frac {\sqrt{\kappa\left(\trainP_{k}^{-1}\trainK\right)} - 1}{\sqrt{\kappa\left(\trainP_{k}^{-1}\trainK\right)} + 1} \right)^{p} \left\Vert \bu^* - \bu_{0} \right\Vert_{\trainK}.
\end{equation}
Our goal is therefore to bound the condition number of $(L_k L_k^\top + \sigma^2 I)^{-1} (K_{XX} + \sigma^2 I)$.
To do so, we will first show that $L_k L_k^\top$ rapidly converges to $K_{XX}$ as the rank $k$ increases.
We begin by restating the primary convergence result of \cite{harbrecht2012low}:
\begin{lemma}[\citet{harbrecht2012low}]
\label{thm:harbrecht}
  If the eigenvalues of a positive definite matrix $K_{XX} \in \reals^{n \times n}$ satisfy $4^{k}\lambda_{k} \lesssim \exp(-bk)$ for some $b>0$, then
  the rank $k$ pivoted Cholesky decomposition $L_{k} L_k^\top$ satisfies
  $$
    \textrm{Tr}(K_{XX} -  L_k L_k^\top) \lesssim n\exp(-bk).
  $$
\end{lemma}
(See \citet{harbrecht2012low} for proof.)
Intuitively, if the eigenvalues of a matrix decay very quickly (exponentially), then it is very easy to approximate with a low rank matrix, and the pivoted Cholesky algorithm rapidly constructs such a matrix.
While there has been an enormous amount of work understanding the eigenvalue distributions of kernel functions (e.g., \cite{wathen2015spectral}), in this paper we prove the following useful bound on the eigenvalue distribution of univariate RBF kernel matrices:
\begin{lemma}
\label{thm:eigenvalue_bound}
Given $x_1, \ldots, x_n \in [0, 1]$, the univariate RBF kernel matrix $K_{XX} \in \mathbb{R}^{n \times n}$ with $K_{ij} = \exp \left(-\gamma(x_i - x_j)^{2}\right)$ has eigenvalues bounded by:
\begin{equation*}
  \lambda_{2l+1} \leq
  2n e^{-\gamma/4} I_{l+1}(\gamma/4) \sim
  \frac{2n e^{-\gamma/4}}{\sqrt{\pi\gamma}}
  \left( \frac{e\gamma}{8(l+1)} \right)^{l+1}
\end{equation*}
where $I_j$ denotes the modified Bessel function of the first kind with parameter $j$.
\end{lemma}
(See \autoref{sec:proofs} for proof.)
Thus, the eigenvalues of an RBF kernel matrix $K_{XX}$ decay \emph{super-exponentially}, and so the bound given by \autoref{thm:harbrecht} applies.

\autoref{thm:eigenvalue_bound} lets us argue for the pivoted Cholesky decomposition as a preconditioner.
%To bound the condition number of the matrix $\trainP^{-1} \trainK$, we begin bounding the difference between a linear solve with the $\trainP$ and the full matrix $\trainK$.
%
%\begin{theorem}
%Let $\by$ be any vector, and let $\bu^{*} = \trainK^{-1}\by$ and $\tilde{\bu} = \trainP_{k}^{-1}\by$, where $\trainP_{k}=L_{k} L_K^\top + \sigma^{2}I$ and $L_{k} L_k^\top$ is the rank $k$ pivoted Cholesky decomposition of $K_{XX}$.
%There exists some $b>0$ so that
%$$
  %\frac{\Vert \bu^{*} - \tilde{\bu}\Vert_{2}}{\Vert\bu^{*}\Vert_{2}} \lesssim \frac{n\exp(-bk)}{\sigma}.
%$$
%\end{theorem}
%
Intuitively, this theorem states that the pivoted Cholesky $L_k L_k$ converges rapidly to $K_{XX}$.
Alternatively, the preconditioner $(L_k L_k^\top + \sigma^2 I)^{-1}$ converges rapidly to $\trainK^{-1} = (K_{XX} + \sigma^2 I)^{-1}$ -- the optimal preconditioner in terms of the number of CG iterations.
We explicitly relate \autoref{thm:eigenvalue_bound} to the rate of convergence of CG by bounding the condition number:
\newtheorem*{thm:condition_number}{\autoref{thm:condition_number} (Restated)}
\begin{thm:condition_number}
  Let $K_{XX} \in \reals^{n \times n}$ be a univariate RBF kernel matrix.
  Let $L_{k} L_k^\top$ be the rank $k$ pivoted Cholesky decomposition of $K_{XX}$, and let $\trainP_{k} = L_k L_k^\top + \sigma^{2}I$.
  Then there exists a constant $b>0$ so that the condition number $\kappa(\trainP^{-1}\trainK)$ satisfies the following inequality:
  \begin{equation}
    \kappa \left( \trainP_{k}^{-1}\trainK \right)
    \triangleq \left\Vert \trainP_{k}^{-1}\trainK \right\Vert_{2} \left\Vert \trainK^{-1}\trainP_{k} \right\Vert_{2}
    \leq \left( 1 + \bigo{n\exp(-bk)} \right)^2.
  \end{equation}
\end{thm:condition_number}
(See \autoref{sec:proofs} for proof.)
\autoref{thm:condition_number} lets us directly speak about the impact of the pivoted Cholesky preconditioner on CG convergence.
Plugging \autoref{thm:condition_number} into standard CG convergence bound \eqref{eq:condition_number_convergence}:
\begin{align*}
  \left\Vert \bu^* - \bu_{k} \right\Vert_{\trainK}
  &\leq 2 \left( \frac{\sqrt{\kappa\left(\trainP_{k}^{-1}\trainK\right)} - 1}{\sqrt{\kappa\left(\trainP_{k}^{-1}\trainK\right)} + 1} \right)^{p} \left\Vert \bu^* - \bu_{0} \right\Vert_{\trainK}
  \\
  &\leq 2 \left( \frac{{{1 + \bigo{n\exp(-bk)}} - 1}} { {{1 + \bigo{n\exp(-bk)}} + 1}} \right)^{p} \left\Vert \bu^* - \bu_{0} \right\Vert_{\trainK}
  \\
  &= 2 \left(\frac{1}{1 + \bigo{\exp(kb) / n }}\right)^{p} \left\Vert \bu^{*} - \bu_{0} \right\Vert_{\trainK}.
\end{align*}
\end{proof}

%!tex root=../main.tex
\section{Proofs of Lemmas}
\label{sec:proofs}

\subsection{Proof of \autoref{thm:condition_number}}

\begin{proof}
  Let $K_{XX} \in \reals^{n \times n}$ be a univariate RBF kernel matrix, and let $L_{k} L_k^\top$ be its rank $k$ pivoted Cholesky decomposition.
  Let $E$ be the difference between $K_{XX}$ and its low-rank pivoted Cholesky approximation -- i.e. $E = K_{XX} - L_k L_k^\top$.
  We have:
  \begin{align*}
    \kappa \left( \trainP_{k}^{-1}\trainK \right)
    &\triangleq \left\Vert \trainP_{k}^{-1}\trainK \right\Vert_{2} \left\Vert \trainK^{-1}\trainP_{k} \right\Vert_{2}
    \\
    &= \left\Vert \left( L_k L_k^\top + \sigma^2 I \right)^{-1} \left(K_{XX} + \sigma^2 I \right) \right\Vert_{2}
    \left\Vert \left( L_k L_k^\top + \sigma^2 I \right) \left(K_{XX} + \sigma^2 I \right)^{-1} \right\Vert_{2}
    \\
    &= \left\Vert \left( L_k L_k^\top + \sigma^2 I \right)^{-1} \left(L_k L_K^\top + E + \sigma^2 I \right) \right\Vert_{2}
    \left\Vert \left( K_{XX} - E + \sigma^2 I \right) \left(K_{XX} + \sigma^2 I \right)^{-1} \right\Vert_{2}
    \\
    &= \left\Vert I + \left( L_k L_k^\top + \sigma^2 I \right)^{-1} E \right\Vert_{2}
    \left\Vert I - \left(K_{XX} + \sigma^2 I \right)^{-1} E \right\Vert_{2}
  \end{align*}
  Applying Cauchy-Schwarz and the triangle inequality we have
  \begin{align*}
    \kappa \left( \trainP_{k}^{-1}\trainK \right)
    &\leq \left( 1 + \left\Vert \left( L_k L_k^\top + \sigma^2 I \right)^{-1} \right\Vert_2 \left\Vert E \right\Vert_{2} \right)
      \left( 1 + \left\Vert \left( K_{XX} + \sigma^2 I \right)^{-1} \right\Vert_2 \left\Vert E \right\Vert_{2} \right)
  \end{align*}
  Let $c$ be some constant such that $c \geq \left\Vert \left( L_k L_k^\top + \sigma^2 I \right)^{-1} \right\Vert_{2}$
  and $c \geq \left\Vert \left( K_{XX} + \sigma^2 I \right)^{-1} \right\Vert_{2}$. Then:
  \begin{align*}
    \kappa \left( \trainP_{k}^{-1}\trainK \right)
    &\leq \left( 1 + c \left\Vert E \right\Vert_{2} \right)^2
  \end{align*}
  \citet{harbrecht2012low} show that $E$ is guaranteed to be positive semi-definite, and therefore $\Vert E \Vert_2 \leq \tr{E}$.
  Recall from \autoref{thm:harbrecht} and \autoref{thm:eigenvalue_bound} that $\tr{E} = \tr{K_{XX} - L_k L_k^\top} \lesssim n\exp(-bk)$ for some $b > 0$.
  Therefore:
  \begin{align*}
    \kappa \left( \trainP_{k}^{-1}\trainK \right)
    &\leq \left(1 + \bigo{n\exp(-bk)}\right)^2.
  \end{align*}
\end{proof}

\subsection{Proof of \autoref{thm:eigenvalue_bound}}
\begin{proof}
  We organize the proof into a series of lemmata.  First, we observe
  that if there is a degree $d$ polynomial that approximates
  $\exp(-\gamma r^2)$ to within some $\epsilon$ on $[-1,1]$,
  then $\lambda_{d+1}(K_{XX}) \leq n\epsilon$
  (\autoref{lemma:interpbound}).
  Then in \autoref{lemma:errbnd}, we show that if $p_l$ is a
  truncated Chebyshev expansions of degree $2l$, then
  $|p_l(r)-\exp(-\gamma r^2)| < 2 e^{-\gamma/4} I_{l+1}(\gamma/4)$;
  the argument involves a fact about sums of modified Bessel functions
  which we prove in \autoref{lemma:Isum}.
  Combining these two lemmas yields the theorem.
\end{proof}
\begin{lemma}\label{lemma:interpbound}
  Given nodes $x_1, \ldots, x_n \in [0,1]$, define the kernel matrix
  $K \in \reals^{n \times n}$ with $k_{ij} = \phi(x_i-x_j)$.  Suppose
  the degree $d$ polynomial $q$ satisfies $|\phi(r)-q(r)| \leq
  \epsilon$ for $|r| \leq 1$.  Then
  \[
    \lambda_{d+1}(K) \leq n \epsilon.
  \]
\end{lemma}
\begin{proof}
  Define $\tilde{K} \in \reals^{n \times n}$ with $\tilde{k}_{ij} = q(x_i-x_j)$.
  Each column is a sampling at the $X$ grid of a $\deg(q)$ polynomial, so
  $\tilde{K}$ has rank at most $\deg(q)$.  The entries of the
  error matrix $E = K-\tilde{K}$ are bounded in magnitude by
  $\epsilon$, so $\|E\|_2 \leq n\epsilon$ (e.g.~by Gershgorin's circle theorem).
  Thus, $\lambda_{d+1}(K) \leq \lambda_{d+1}(\tilde{K}) + \|E\|_2 = n\epsilon$.
\end{proof}

\begin{lemma}~\label{lemma:errbnd}
  For $x \in [-1,1]$,
  \[
    |\exp(-\gamma x^2)-p_l(x)| \leq 2 e^{-\gamma/4} I_{l+1}(\gamma/4).
  \]
\end{lemma}
\begin{proof}
Given that
$|(-1)^j T_{2j}(x)| \leq 1$ for any $x \in [-1,1]$, the tail
admits the bound
\[
  |\exp(-\gamma x^2)-p_l(x)| \leq
  2 e^{-\gamma/2} \sum_{j=l+1} I_j(\gamma/2).
\]
Another computation (\autoref{lemma:Isum}) bounds the sum of the
modified Bessel functions to yield
\[
  |\exp(-\gamma x^2)-p_l(x)| \leq
  2 e^{-\gamma/4} I_{l+1}(\gamma/4).
\]
\end{proof}

\begin{lemma}\label{lemma:Isum}
  \[
    \sum_{j=l+1}^\infty I_j(\eta) \leq \exp(\eta/2) I_{l+1}(\eta/2)
  \]
\end{lemma}
\begin{proof}
Take the power series expansion
\[
  I_j(\eta) =
    \sum_{m=0}^\infty \frac{1}{m! (m+j)!} \left(\frac{\eta}{2}\right)^{2m+j}
\]
and substitute to obtain
\[
  \sum_{j=l+1}^\infty I_j(\eta) =
    \sum_{j=l+1}^\infty \sum_{m=0}^\infty
    \frac{1}{m! (m+j)!} \left(\frac{\eta}{2}\right)^{2m+j}.
\]
All sums involved converge absolutely, and so we may reorder to obtain
\[
  \sum_{j=l+1}^\infty I_j(\eta) =
  \sum_{m=0}^\infty \frac{1}{m!} \left(\frac{\eta}{2}\right)^m
  \sum_{j=l+1}^\infty
    \frac{1}{(m+j)!} \left(\frac{\eta}{2}\right)^{m+j}.
\]
Because it is the tail of a series expansion for the exponential, we
can rewrite the inner sum as
\[
  \sum_{j=l+1}^\infty
  \frac{1}{(m+j)!} \left(\frac{\eta}{2}\right)^{m+j} =
  \frac{\exp(\xi_m/2)}{(m+l+1)!} \left(\frac{\xi_m}{2}\right)^{m+l+1}
\]
for some $\xi_m$ in $[0,\eta]$, and thus
\[
  \sum_{j=l+1}^\infty
  \frac{1}{(m+j)!} \left(\frac{\eta}{2}\right)^{m+j} \leq
  \frac{\exp(\eta/2)}{(m+l+1)!} \left(\frac{\eta}{2}\right)^{m+l+1}.
\]
Substituting into the previous expression gives
\begin{align*}
  \sum_{j=l+1}^\infty I_j(\eta)
  &\leq
  \sum_{m=0}^\infty \frac{1}{m!} \left(\frac{\eta}{2}\right)^m
  \frac{\exp(\eta/2)}{(m+l+1)!} \left(\frac{\eta}{2}\right)^{m+l+1} \\
  &=
  \exp\left(\frac{\eta}{2}\right)
  \sum_{m=0}^\infty \frac{1}{m! (m+l+1)!}
  \left(\frac{\eta}{2}\right)^{2m+l+1} \\
  &=
  \exp(\eta/2) I_{l+1}(\eta/2).
\end{align*}
\end{proof}

\end{document}